\documentclass[journal,draftcls,onecolumn]{IEEEtran}

\usepackage{ifpdf}
\usepackage{cite}

\usepackage{graphicx}

\usepackage{amsmath}
\usepackage{tabularx}
\usepackage{array}
\usepackage{fixltx2e}
\usepackage{stfloats}
\usepackage{url}
\usepackage{pgfplots}
\usepackage{float}
\usepackage{caption}
\usepackage{algorithm} 
\usepackage{algpseudocode}
\usepackage{subcaption}
\usepackage{tikz}
\usepackage{bm}
\usepackage{amssymb}  
\usepackage{mathtools}  
\usepackage{comment}
\usepackage{forest}

\usetikzlibrary{shapes,arrows,positioning,fit}

\newcommand{\norm}[1]{\left\lVert #1 \right\rVert}
\newcommand{\inner}[2]{\left\langle #1, #2 \right\rangle}

\def\FVmaxa{21}
\def\FVmax{20} 
\def\Ftzero{7} 
\def\Ftone{12} 
\def\Fttwo{10} 
\def\Ftthree{6} 
\def\Ftfour{4} 
\def\Famax{1.6666} 
\def\Fdmax{3.333} 

\begin{document}
	
	\newcommand{\define}[1]{\textbf{#1}}
	
	\newcommand{\fmaw}[1]{{\textbf{[FIX-AW!]}}\footnote{{\textbf{[AW]:}}#1}}
	\newcommand{\fmsw}[1]{{\textbf{[FIX-SW!]}}\footnote{{\textbf{[SW]:}}#1}}
	\newcommand{\fmbw}[1]{{\textbf{[FIX-BW!]}}\footnote{{\textbf{[BW]:}}#1}}
	
	%
	\title{A Time-efficient Prioritised Scheduling Algorithm to Optimise Initial Flock Formation of Drones}
	%
	%
	%
	
	\author{Sujan~Warnakulasooriya,
		Andreas~Willig
		and~Xiaobing~Wu
		\thanks{Sujan Warnakulasooriya is with the Wireless Research Centre, University of Canterbury (UC), Christchurch 8140, New Zealand, and also with the Computer Science and Software Engineering Department,  University of Canterbury (UC), Christchurch 8140, New Zealand
			(e-mail: sujan.warnakulasooirya@pg.canterbury.ac.nz).}
		\thanks{Andreas Willig is with the Computer Science and Software Engineering Department, University of Canterbury (UC), Christchurch 8140, New Zealand
			(e-mail: andreas.willig@canterbury.ac.nz).}
		\thanks{Xiaobing Wu is with the Wireless Research Centre, University of Canterbury (UC), Christchurch 8140, New Zealand (e-mail: barry.wu@canterbury.ac.nz).}
		\thanks{Manuscript received Xxx xx, xxxx; revised Xxx xx, xxxx.}}

\maketitle

\begin{abstract}
	Drone applications continue to expand across various domains, with flocking offering enhanced cooperative capabilities but introducing significant challenges during initial formation. Existing flocking algorithms often struggle with efficiency and scalability, particularly when potential collisions force drones into suboptimal trajectories. This paper presents a time-efficient prioritised scheduling algorithm that improves the initial formation process of drone flocks. The method assigns each drone a priority based on its number of potential collisions and its likelihood of reaching its target position without permanently obstructing other drones. Using this hierarchy, each drone computes an appropriate delay to ensure a collision-free path. Simulation results show that the proposed algorithm successfully generates collision-free trajectories for flocks of up to 5000 drones and outperforms the coupling-degree-based heuristic prioritised planning method (CDH-PP) in both performance and computational efficiency.  
\end{abstract}


%
\IEEEpeerreviewmaketitle

\section{Introduction}
\label{sec-intro} 

\IEEEPARstart{O}{ver} the past couple of decades, unmanned aerial vehicles (UAVs), or drones, have experienced rapid growth and evolved into diversified applications across multiple domains \cite{Mohsan_2022}, resulting in drones becoming more accessible and highly capable over the past few years \cite{Ahmed_2022}. Interactive flying systems have been identified in 16 different domains and over 100 applications by observing the latest research \cite{Herdel_2022}. However, drones face limitations in flight duration due to battery capacity and payload threshold when performing complex tasks \cite{Liu_2020}. One solution to mitigate these limitations is to allow multiple drones to operate together as a flock \cite{Alotaibi_2023}. The coordinated and cohesive movement inside a defined environment with rigid relative positions, inspired by animals such as birds or fish, is defined as flocking \cite{Onur_2024}. Flocking behaviour offers advantages such as added flexibility, divided workload, faster completion of tasks, redundancy in case of an individual drone failure, and better overall coverage \cite{Roldan_2016}, while introducing the cost of increased complexity in orchestrating an entire group of drones. Applications in agriculture \cite{Ju_2020}, delivery systems \cite{Xijing_2022}, \cite{Alkouz_2023}, environmental monitoring \cite{Brust_2015}, infrastructure maintenance \cite{Uzakov_2020}, landmine detection \cite{Cimino_2015}, search and rescue \cite{Amala_2023}, and surveillance \cite{Nigam_2012} are among the examples of enhanced drone flocking capabilities. \par

While appreciating the enhanced capabilities of drone flocking, several new challenges arise when multiple drones attempt to operate in the same geographical area. A drone flock faces interconnected challenges including control system design, NP-hard path planning, implementation of collision avoidance protocols, communication network reliability, continuous monitoring requirements, and computational scalability as swarm sizes increase \cite{Iqbal_2022}. Vasarhelyi et al. \cite{Vasarhelyi_2018} successfully demonstrated that flocking models require explicit handling of constrained motion, communication delays, and barriers, resulting in additional model complexity and increased tunable parameters that necessitate an evolutionary optimisation framework when they validated the seamless movement of a flock of 30 drones autonomously.  In this paper, the initial formation of a flock of drones is investigated. This is considered as an initial flock formation problem where the drones move from starting positions to their respective target positions within the flock. This problem can be essentially categorised as a drone path planning problem. \par

An efficient path-planning algorithm should provide a solution that is both complete and optimal \cite{Jones_2023}. As noted by Gasparetto et al. \cite{Gasparetto_2015}, path planning is concerned solely with the geometric aspect of motion and does not account for time. In contrast, trajectory planning enriches a geometric path by specifying its temporal profile, defining how position, velocity, and acceleration evolve along the path. When operating as a flock and planning individual paths, drones must minimise formation time and total travel distance, ensure collision-free travel, and tolerate sensing and navigation errors to an acceptable degree, all while maintaining reasonable computation times \cite{Aggarwal_2020}. Various path planning or trajectory planning approaches have been proposed in the literature to address the problem of initial flock formation for a fixed number of $n$ drones in an obstacle-free environment, starting from known arbitrary starting positions and moving to designated target positions in a desired flock geometry, without collisions and in the shortest possible time. Existing approaches emphasise accurate and reliable flock formation—whether through geometric path planning (e.g., cell decomposition, rapidly-exploring random trees (RRTs), artificial potential fields) or trajectory‑based control (e.g., consensus laws, ant colony optimisation, heuristic scheduling, space–time graph pruning). However, these methods often compromise efficiency in flocking time and travel distance, relying on reactive avoidance, simplified environments, or limited scalability. This paper introduces a novel approach, time-efficient prioritised scheduling (TPS), designed to coordinate a flock of drones to form their initial formation without any collisions, even at large scales. The TPS algorithm assumes each drone travels along a straight-line path from its starting position to its destination. To prevent inter-drone collisions, the trajectory of each drone is modified by introducing a calculated starting delay. Even though the individual paths are fixed for all the drones, the proposed approach enables each drone to move from a unique starting location to a unique target location while avoiding inter-drone collisions. The primary contribution of this work is a prioritised scheduling algorithm that guarantees collision‑free trajectories while minimising travel distances. Compared with the coupling-degree-based heuristic prioritized planning method (CDH-PP), the proposed TPS algorithm achieves collision-free travel from every drone   and it scales up to 5,000 drones in the simulations. \par

The remainder of this paper is organised as follows: Section~\ref{sec-related} discusses related work on flock path planning and trajectory planning, and Section~\ref{sec-problem2} formulates the initial flock formation problem. Section~\ref{sec-scheduling}  presents the proposed time-efficient prioritised scheduling (TPS) algorithm.  Section~\ref{sec-results} presents the simulation results. 
 Finally, the conclusion and future directions are outlined in Section~\ref{sec-conclusion}.

\section{Related work} 
\label{sec-related}
Initial drone flock formation can be addressed through either path planning, which focuses on geometric routes, or trajectory planning, which enriches these routes with timing and velocity constraints. Numerous studies have explored both perspectives, proposing methods to tackle challenges such as obstacle avoidance, collision prevention, and formation maintenance. Yet, most approaches emphasise overall swarm performance rather than the unique difficulties of initial formation, where drones depart from random stationary positions and must converge into a desired geometry. Existing approaches can therefore be grouped into path‑planning methods and trajectory‑planning methods. Their performance in the context of initial flock formation is analysed below. \par

\subsection{Path Planning Methods} 

Path planning for drone flocks has been addressed through various algorithmic approaches, with cell decomposition methods being particularly prominent. Iswanto et al. \cite{Iswanto_2017} proposed a modified cell decomposition algorithm that combined fuzzy logic, cell decomposition, and potential field algorithms to enable a formation pattern for three drones. This approach divides the operating space into cells and incorporates potential field weighting to find shorter paths in an environment with static and dynamic obstacles. However, cell decomposition creates grids, leading to suboptimal paths that follow grid boundaries rather than true optimal trajectories. In addition, local decision-making does not consider global path optimality across the entire formation and tends to wait for drone formation completion rather than continuing toward goals. Hence, in an initial flock formation scenario, the modified cell decomposition algorithm compromises on path optimality by local decision-making and prioritising formation completion. \par

Research on RRT-based path planning for UAVs has focused on addressing the challenges of generating optimal paths in complex environments. Kaur and Prasad \cite{Kaur_2019} proposed a Rapidly-exploring Random Tree (RRT) algorithm for optimal path planning in a flock of up to five autonomous drones. Each drone uses RRT with different step sizes, and once a path is generated, the tree vertices are deleted to prevent other drones from using them. This algorithm improved consistency by generating similar paths across iterations and achieving minimum distances between nodes in a 2D configuration space. Their approach included a path reconstruction strategy specifically designed for handling multiple UAVs. The sequential deletion of tree vertices creates increasingly constrained search spaces and impacts fairness for subsequent UAVs, leading to progressively worse paths. Despite these limitations, RRT offers a balance between execution time and overall path optimisation for drone applications \cite{Cai_2022}. \par

Li and Fang \cite{Shuang_Li_2021} developed an algorithm inspired by pigeon swarm behaviour to help drones adjust their formation in response to environmental changes. The proposed algorithm employs a modified pigeon behaviour algorithm for local visual field obstacle avoidance and a hierarchical early warning mechanism for inter-drone collision avoidance, and simulations were conducted with up to nine drones in a flock. This algorithm mainly focuses on collision and obstacle avoidance, and path planning can be considered as a purely reactive mechanism with no path optimisation. Individual drones often make greedy local decisions based on immediate observations to avoid collisions. While effective in the short term, such decisions lead to global path inefficiencies and extended formation times. In addition, forcing specific formations regardless of whether they're beneficial for the current scenario adds unnecessary path length. \par

Recent research has focused on artificial potential field (APF) methods for drone swarm path planning and formation control. Zhang et al. \cite{Zhang_2022} developed a hybrid approach combining APF with a novel GA-DPSO evolutionary algorithm for UAV formation transformation, demonstrating effectiveness in optimising paths for multiple drones while addressing battery capacity constraints. When calculating paths for 800 drones in a drone art light show, completion times were found to be significantly reduced compared with traditional path planning methods. However, while all drones reached their target simultaneously, the varying travel distances meant some drones did not take the shortest path. The theoretical minimum time for formation completion was not provided, leaving some uncertainty about the optimality of the formation time. \par

\subsection{Trajectory Planning Methods} 

Sabetghadam et al. \cite{Sabetghadam_2022} developed a distributed algorithm using Voronoi partitioning that enables up to 100 drones to generate collision-free trajectories through receding horizon optimisation, where each drone solves individual optimisation problems using local constraints derived from Voronoi space partitioning. Individual drones begin their journeys from randomly selected starting positions toward randomly chosen target positions. During flight, they share location data with nearby drones in their vicinity. This communication enables each drone to refresh its designated Voronoi region, compute cost calculations, establish operational limitations, and resolve individual optimisation challenges. While this approach may not consistently identify the most direct route possible, it substantially decreases computational processing time when compared to the Buffered Voronoi Cells (BVC) method. When using this method, each replanning cycle generates locally optimal paths without considering global trajectory optimisation that fails to capture collective efficiency. The distributed nature prevents globally optimal formation paths, as each drone only considers its immediate Voronoi cell, which leads to longer travel distances as drones could be directed away from their goals to avoid collisions. Constant replanning also adds a significant computational burden as the number of drones in the flock increases. Moreover, a robust sensing and communication mechanism is essential to the success of this algorithm. \par
 
Li et al. \cite{Hui_Li_2019} proposed a coupling-degree-based heuristic prioritised planning method (CDH-PP) to generate paths for a drone swarm containing up to 20 drones by decomposing the problem into sequential single-drone planning tasks. Initially, paths were planned using the anytime repairing sparse A* algorithm (AR-SAS), ignoring potential collisions. Then, a coupling degree matrix calculates the collision relationships among drones using the initial paths and assigns a priority to each drone based on the number of potential collisions. Subsequently, a replanning of paths is carried out sequentially according to the priority, and each lower prioritised drone must find a new, collision-free path by exploring different route options. This approach to prioritised planning inherently produces suboptimal solutions for lower prioritised drones. In addition, the initially planned path only serves as a baseline to calculate the coupling degree matrix and adds little value to the final paths. The time limit of 0.5s per path limits the possibility of improving the current path, while scalability becomes limited as the number of drones increases in the flock, which could affect the success rate of task completion as well. \par

Recent research has explored ant colony optimisation (ACO) applications for multi-agent path planning, particularly in drone formations. Suzuki et al. \cite{Suzuki_2022} developed an ACO-based method for generating fair paths in consecutive pattern formations, emphasising equitable travel distances among agents rather than solely minimising total distance to prevent battery exhaustion and extend performance time. Their approach enabled agents to execute approximately 20\% more formation patterns without collisions compared to conventional ACO methods. In simulation, paths for up to 20 drones were calculated, guided by a fitness function that adaptively balances weights to promote stable motion. The paths are computed in discrete steps, where each drone selects its next node while preventing collisions by avoiding node overlap. This method intentionally sacrifices path optimality for fairness, while collision avoidance further contributes to this suboptimality. \par

A leader–follower formation control strategy was introduced by Mukherjee and Namuduri \cite{Mukherjee_2019}, integrating consensus rules with social potential functions to maintain collision avoidance and network connectivity among five drones. The leader drone, guided by a corrective force proportional to its position error, directs the group toward the target. Each follower determines its control input from its position and velocity deviations relative to designated neighbours, structured by a predefined geometric matrix. Although the approach is effective once the formation is established, its reliance on neighbour interactions means some drones may initially travel longer paths, as their movement is influenced more by neighbouring positions than by direct progress toward the destination. \par

The Sequential Arrival method proposed by Babel \cite{Babel_2019} achieves a time-shift and path-extension process for individual drones when all of them arrive at a common landing strip. This required delay is achieved by employing a path adjustment method, where individual drone paths are extended to match their required arrival time. While adjusting paths, the algorithm also checks for conflicts between drones and chooses alternative waypoints if safety distances are violated. In this research, scheduling is used as a tool to effectively utilise a common destination, but not as a tool to avoid collisions.  \par

When a potential collision is detected, Bahabry et al. \cite{Bahabry_2019} propose a proactive scheduling mechanism to avoid collisions within a fleet of drones tasked with covering a set of known events, while minimising energy usage. Drones are sorted from highest to lowest stored energy, and tasks are assigned to each drone using a mixed integer linear programme (MILP), aiming to maximise the number of events a drone can cover whilst minimising total energy consumption. Paths are subsequently planned using Dijkstra’s algorithm to determine the shortest routes between events and charging stations. Drones are then assigned paths sequentially, and if a collision risk is identified, the affected drone is instructed to hover at a safe location for a delay period. Hovering can be considered a mid-position delay; however, differences in the number of tasks assigned to individual drones may introduce inefficiencies within the system. \par

When solving a similar research problem Papa et al. \cite{Papa_2020} introduces a reactive scheduling mechanism to cater on-demand drone parcel delivery that accounts for energy constraints and collision avoidance within a predefined aerial highway system. Delivery requests are processed individually, and paths are computed using a multi-source A* algorithm with a branching factor of one, ensuring scalability and real-time responsiveness. The algorithm models battery limitations by allowing drones to recharge once before and once after client delivery, and enforces collision-free trajectories by pruning conflicting edges in a space-time graph based on safety distance requirements. If a path segment is unavailable due to potential conflict, the drone is instructed to pause at its current location for a time interval. While this approach ensures safe and energy-aware routing, the lack of task batching and the single-path search strategy may lead to inefficiencies in high-demand scenarios. \par

In summary, previous studies have advanced drone flocking through both path‑planning methods, which emphasise geometric optimisation, and trajectory‑planning methods, which incorporate timing and velocity constraints. While these approaches achieve accurate and reliable formations, they involve trade‑offs: some extend travel distances to avoid collisions, others prolong formation time to ensure fairness, and many assume simplified environments that limit scalability. In the context of initial flock formation, these limitations become more pronounced, as drones must depart from random stationary positions and converge efficiently into a desired geometry. To address these gaps, this study introduces a time‑efficient prioritised scheduling (TPS) algorithm that guarantees collision‑free trajectories along straight‑line paths by introducing a calculated and minimised starting delays to drones that require a delay to avoid a collision. By constraining paths and excluding obstacles, TPS reduces computational overhead and scales effectively to swarms of over 5,000 drones, improving both formation time and overall efficiency during the critical initial phase.

\section{Problem description}
\label{sec-problem2} 

\begin{figure}[h]
	\centering
	\includegraphics[width=0.65\textwidth]{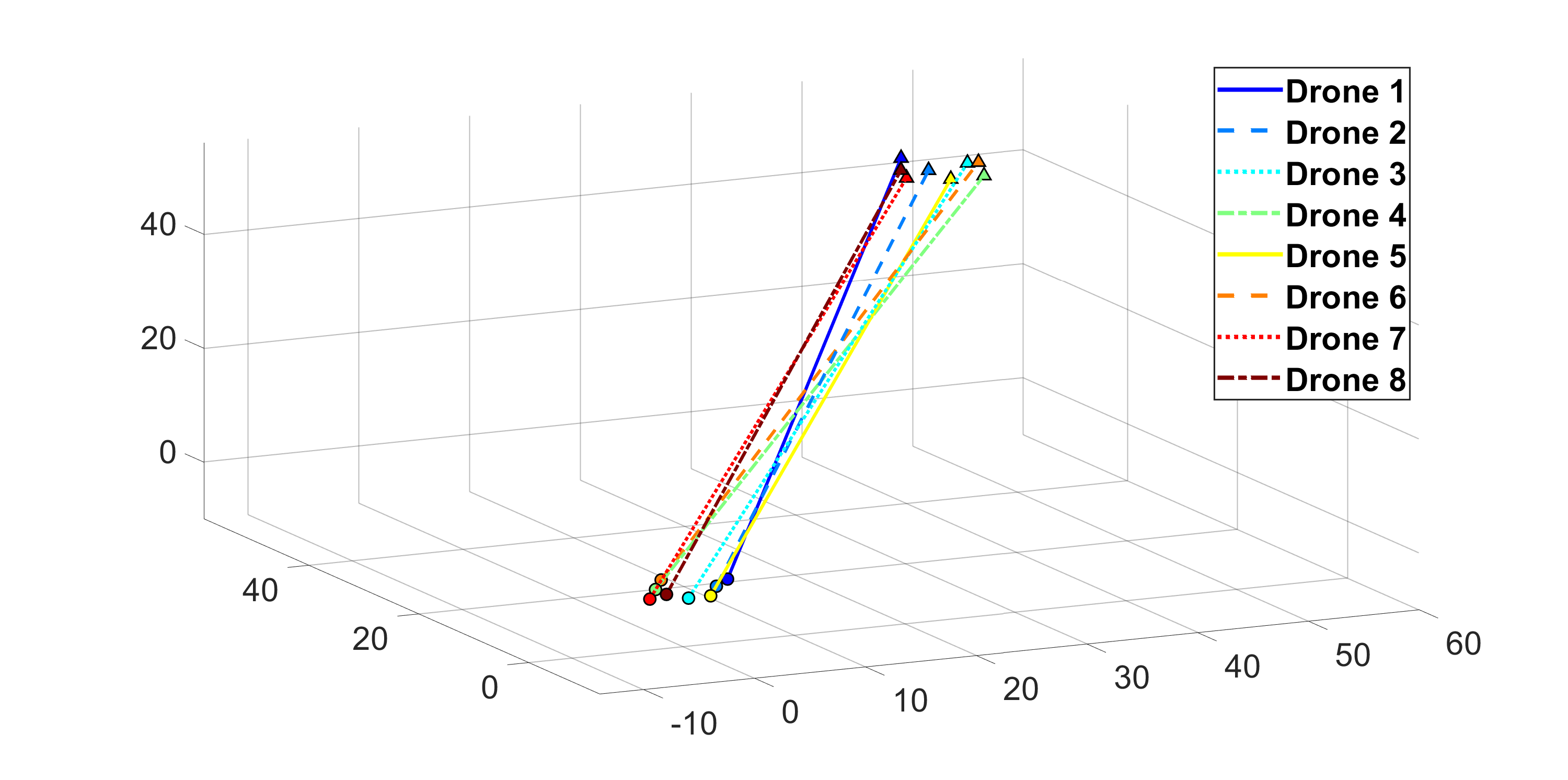}
	\caption{Example travel path for 8 drones }
	\label{fig:scenario}
\end{figure}

Assume a scenario where there are a fixed number of drones $n$ operating within a designated three-dimensional (3D) space without any obstacles. Each drone $i$ operates from a unique starting position $S^i_\text{str}$ to a corresponding unique target position $S^i_\text{tgt}$ both of which are known in advance and arbitrarily located. Uniform physical constraints—namely, maximum acceleration, maximum velocity, and maximum deceleration govern all drones, and their individual trajectories are determined accordingly. As the environment contains no obstacles, drones move in straight lines between their starting and target positions. The flocking time is defined as the duration from the moment when the first drone begins moving until the moment when the last drone comes to a complete stop. The main objective of the problem is to minimise the flocking time of the drones. An example with paths for 8 drones is shown in Fig.~\ref{fig:scenario}. The circles represent the starting positions, and the triangles represent the target positions.

\subsection{Drone Dynamics and Paths}

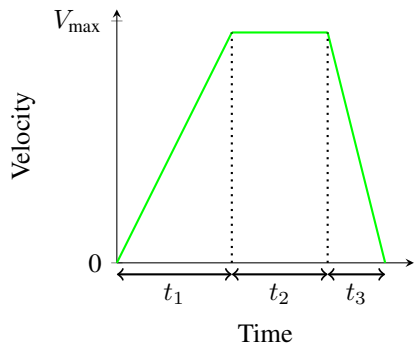
\begin{figure}[htbp]
	\centering
	\begin{tikzpicture}
		\begin{axis}[
			width=0.335\textwidth, 
			height=0.3\textwidth, 
			axis lines=left,
			xlabel={Time},
			ylabel={Velocity},
			xlabel style={at={(axis description cs:0.5,-0.05)},anchor=north}, 
			ylabel near ticks,
			xtick=\empty,
			ytick={0, \FVmaxa},
			yticklabels={0, $V_{\text{max}}$},
			clip=false,
			ymin=0,
			ymax=\FVmaxa+1,
			xmin=0,
			xmax=\Ftone+\Fttwo+\Ftthree+3,
			domain=0:\Ftone+\Fttwo+\Ftthree
			]
			
			\addplot[green, thick] expression[domain=0:\Ftone]{\Famax*x};
			
			\addplot[green, thick] expression[domain=\Ftone:\Ftone+\Fttwo]{\FVmax};
			
			\addplot[green, thick] expression[domain=\Ftone+\Fttwo:\Ftone+\Fttwo+\Ftthree]{\FVmax-\Fdmax*(x-\Ftone-\Fttwo)};
			
			\draw[dotted, thick] (axis cs:\Ftone,0) -- (axis cs:\Ftone,\FVmax);
			\draw[dotted, thick] (axis cs:\Ftone+\Fttwo,0) -- (axis cs:\Ftone+\Fttwo,\FVmax);
			
			\draw[<->, thick] (axis cs:0, -1) -- node[below] {$t_1$} (axis cs:\Ftone, -1);
			
			\draw[<->, thick] (axis cs:\Ftone, -1) -- node[below] {$t_2$} (axis cs:\Ftone+\Fttwo, -1);
			
			\draw[<->, thick] (axis cs:\Ftone+\Fttwo, -1) -- node[below] {$t_3$} (axis cs:\Ftone+\Fttwo+\Ftthree, -1);
			
		\end{axis}
	\end{tikzpicture}
	\vspace{-0.5em} 
	\caption{Velocity vs. time graph for a drone travelling in a straight line}
	\label{fig:generic VT basic diagram}
\end{figure}

As previously discussed, each drone $i$ follows a straight-line trajectory subject to predefined physical constraints, as depicted in the velocity-time diagram in Fig. \ref{fig:generic VT basic diagram}. It is assumed that the drone undergoes acceleration at a rate of $a^i$ over a duration of $t_1$, reaching its peak velocity, $V^i$. This velocity is sustained for a time period of $t_2$, after which the drone decelerates at a rate of $d^i$ during $t_3$ until the drone to a complete stop. This is considered to be the ideal velocity profile, and by considering the physical constraints, achieves the optimum travel time.

Since drone acceleration ($a^i$), maximum velocity ($V^i$) and deceleration ($a^i$) parameters are fixed, for a given starting ($S^i_\text{str}$) and target ($S^i_\text{tgt}$) position combination of a given drone $i$, travel time of acceleration phase ($t_1$), maximum velocity phase ($t_2$), and deceleration phase ($t_3$) will be constant. In addition, in an event where distance between starting and target positions is small, there might not be sufficient distance for the drone to reach the maximum velocity. In such a scenario, only acceleration phase and deceleration phase exist, and corresponding accelaration and deceleration times should be calculated accordingly.

\subsection{Drone Travel Time and Flocking Time}

Since all drones travel in a straight line with only the physical drone performances limiting its movement, the travel time for the $i$th drone ($t_{\text{travel}}^i$) is given by \eqref{eq:travel_time_drone}, and the travel time depends on the physical constraints of the drone and travel distance only. 

\begin{equation}
	t_{\text{travel}}^i = t_1^i + t_2^i + t_3^i
	\label{eq:travel_time_drone}
\end{equation}

Drone flocking time ($t_{\text{flock}}^n$) is calculated as the duration between the instant the first drone starts its movement and the instant the last drone comes to a halt. The drone that travels the longest distance determines the lower bound of $t_{\text{flock}}^n$, since the flocking time cannot be shorter than the travel time of the drone covering the maximum distance.

\subsection{Constraints}\label{consts} 

Drones will have two key constraints as presented below, \par
(I) All drones travel in a straight line connecting their starting positions to their target positions. This is feasible because no obstacles exist between the starting area and the target area.

(II) Drones have physical limitations in maximum acceleration, velocity and deceleration.

(III) Drones must avoid collisions by maintaining a safe distance from each other at all times. TPS algorithm calculates potential collisions and adds time delays to individual drones ensuring that the collision avoidance criteria are met by keeping the distance between any two drones greater than the collision radius ($R_{\text{col}}$).
\begin{equation}
	\|S^i(t)-S^j(t)\|  > R_{\text{col}} \quad \text{for } i,j \in \{1, 2, \ldots, n\} \text{ and } i \neq j
	\label{eq: collision_detection}
\end{equation}

(IV) The arbitrary starting and target positions satisfy the condition given by \eqref{eq: collision_detection}.

(V) All drones travel in the same general direction.

\subsection{Problem Formulation}

The problem is formulated as developing a strategy that guarantees collision-free flock formation while achieving overall time efficiency. This objective can be formulated using \eqref{eq: prob-form}.

\begin{equation}
	\begin{aligned}
		&\text{Min } t_{\text{flock}}^n \\
		&\text{s.t. Constraints (I)-(V) in \ref{consts}}
	\end{aligned}
	\label{eq: prob-form}
\end{equation}

\section{The Time-efficient Prioritised Scheduling (TPS) method }
\label{sec-scheduling}

As stated in the problem formulation, the objective is to minimise the drone flocking time ($t_{\text{flock}}^n$). On one hand, if all drones start moving immediately, the flocking time may reach its theoretical lower bound; however, such simultaneous movement will inevitably result in collisions. On the other hand, if only one drone moves at a time, the intuitive expectation is that no collisions will occur. Contrary to this intuition, there exist scenarios in which even sequential single-drone movement can lead to collisions, as will be discussed later. Nevertheless, the majority of scenarios can indeed be resolved by allowing only one drone to move at a time, albeit at the cost of extreme inefficiency. The challenge is to identify a middle ground that ensures collision-free trajectories while keeping $t_{\text{flock}}^n$ as close as possible to its theoretical lower bound, achieved by calculating individual start delays for all $n$ drones in the flock. \par

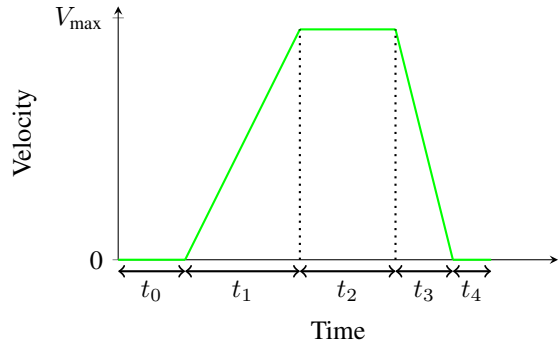
\begin{figure}[htbp]
	\centering
	\begin{tikzpicture}
		\begin{axis}[
			width=0.45\textwidth, 
			height=0.3\textwidth, 
			axis lines=left,
			xlabel={Time},
			ylabel={Velocity},
			xlabel style={at={(axis description cs:0.5,-0.05)},anchor=north}, 
			ylabel near ticks,
			xtick=\empty,
			ytick={0, \FVmaxa},
			yticklabels={0, $V_{\text{max}}$},
			clip=false,
			ymin=0,
			ymax=\FVmaxa+1,
			xmin=0,
			xmax=2*\Ftone+\Fttwo+\Ftzero+\Ftfour+1,
			domain=0:2*\Ftone+\Fttwo+\Ftzero
			]
			
			\addplot[green, thick] expression[domain=0:\Ftzero]{0};

			\addplot[green, thick] expression[domain=\Ftzero:\Ftone+\Ftzero]{\Famax*(x-\Ftzero)};

			\addplot[green, thick] expression[domain=\Ftone+\Ftzero:\Ftone+\Fttwo+\Ftzero]{\FVmax};

			\addplot[green, thick] expression[domain=\Ftone+\Fttwo+\Ftzero:\Ftone+\Fttwo+\Ftzero+\Ftthree]{\FVmax-\Fdmax*(x-\Ftone-\Fttwo-\Ftzero)};

			\addplot[green, thick] expression[domain=\Ftone+\Fttwo+\Ftthree+\Ftzero:\Ftone+\Fttwo+\Ftthree+\Ftzero+\Ftfour]{0};

			\draw[dotted, thick] (axis cs:\Ftone+\Ftzero,0) -- (axis cs:\Ftone+\Ftzero,\FVmax);
			\draw[dotted, thick] (axis cs:\Ftone+\Fttwo+\Ftzero,0) -- (axis cs:\Ftone+\Fttwo+\Ftzero,\FVmax);
			
			\draw[<->, thick] (axis cs:0, -1) -- node[below] {$t_0$} (axis cs:\Ftzero, -1);
			
			\draw[<->, thick] (axis cs:\Ftzero, -1) -- node[below] {$t_1$} (axis cs:\Ftone+\Ftzero, -1);
			
			\draw[<->, thick] (axis cs:\Ftone+\Ftzero, -1) -- node[below] {$t_2$} (axis cs:\Ftone+\Fttwo+\Ftzero, -1);
			
			\draw[<->, thick] (axis cs:\Ftone+\Fttwo+\Ftzero, -1) -- node[below] {$t_3$} (axis cs:\Ftone+\Fttwo+\Ftthree+\Ftzero, -1);
			
			\draw[<->, thick] (axis cs:\Ftone+\Fttwo+\Ftthree+\Ftzero, -1) -- node[below] {$t_4$} (axis cs:\Ftone+\Fttwo+\Ftthree+\Ftzero+\Ftfour, -1);

		\end{axis}
	\end{tikzpicture}
	\vspace{-0.5em} 
	\caption{Velocity vs. time graph for a drone travelling in a straight line with a start delay and end waiting time}
	\label{fig:generic VT diagram}
\end{figure}

\begin{figure}[htbp]
	\centering
	
	\pgfmathsetmacro{\Fsstart}{10}
	\pgfmathsetmacro{\Fsa}{0.5*\Famax*\Ftone*\Ftone}
	\pgfmathsetmacro{\Fsd}{\FVmax*(\Fttwo)+\Fsa}
	\pgfmathsetmacro{\Ftarget}{\Fsd+\FVmax*(\Ftthree)-0.5*\Fdmax*\Ftthree*\Ftthree+\Fsstart}
	
	\begin{tikzpicture}
		\begin{axis}[
			width=0.45\textwidth,
			height=0.4\textwidth, 
			axis lines=left,
			xlabel={Time},
			ylabel={Distance},
			xlabel style={at={(axis description cs:0.5,-0.05)},anchor=north},
			ylabel near ticks,
			xtick=\empty,
			ytick={\Fsstart, \Fsa+\Fsstart, \Fsd+\Fsstart, \Ftarget},
			yticklabels={$S^i_{str}$, $S^i_a$, $S^i_d$, $S^i_{tgt}$},
			clip=false,
			ymin=0,
			xmin=0,
			xmax=2*\Ftone+\Fttwo+\Ftzero+\Ftfour+1,
			ymax= \Ftarget + 20,
			domain=0:2*\Ftone+\Fttwo+\Ftzero
			]
			
			\addplot[green, thick, samples=2, domain=0:\Ftzero]{\Fsstart};
			
			\addplot[green, thick, samples=40, domain=\Ftzero:\Ftone+\Ftzero]{0.5*\Famax*(x-\Ftzero)*(x-\Ftzero)+\Fsstart};
			
			\addplot[green, thick, samples=2, domain=\Ftone+\Ftzero:\Ftone+\Fttwo+\Ftzero]{\FVmax*(x-\Ftone-\Ftzero)+\Fsa+\Fsstart};
			
			\addplot[green, thick, samples=40, domain=\Ftone+\Fttwo+\Ftzero:\Ftone+\Fttwo+\Ftzero+\Ftthree]{\Fsd+\FVmax*(x-\Ftone-\Fttwo-\Ftzero)-0.5*\Fdmax*(x-\Ftone-\Fttwo-\Ftzero)*(x-\Ftone-\Fttwo-\Ftzero)+\Fsstart};
			
			\addplot[green, thick, samples=2, domain=\Ftone+\Fttwo+\Ftthree+\Ftzero:\Ftone+\Fttwo+\Ftthree+\Ftzero+\Ftfour]{\Ftarget};
			
			\draw[dotted, thick] (axis cs:\Ftzero,0) -- (axis cs:\Ftzero,\Fsstart);
			
			\draw[dotted, thick] (axis cs:\Ftone+\Ftzero,0) -- (axis cs:\Ftone+\Ftzero,\Fsa+\Fsstart);
			\draw[dotted, thick] (axis cs:0,\Fsa+\Fsstart) -- (axis cs:\Ftone+\Ftzero,\Fsa+\Fsstart);
			
			\draw[dotted, thick] (axis cs:\Ftone+\Fttwo+\Ftzero,0) -- (axis cs:\Ftone+\Fttwo+\Ftzero,\Fsd+\Fsstart);
			\draw[dotted, thick] (axis cs:0,\Fsd+\Fsstart) -- (axis cs:\Ftone+\Fttwo+\Ftzero,\Fsd+\Fsstart);
			
			\draw[dotted, thick] (axis cs:\Ftone+\Fttwo+\Ftthree+\Ftzero,0) -- (axis cs:\Ftone+\Fttwo+\Ftthree+\Ftzero,\Ftarget);
			\draw[dotted, thick] (axis cs:0,\Ftarget) -- (axis cs:\Ftone+\Fttwo+\Ftthree+\Ftzero,\Ftarget);
			
			\draw[<->, thick] (axis cs:0, -10) -- node[below] {$t_0$} (axis cs:\Ftzero, -10);
			
			\draw[<->, thick] (axis cs:\Ftzero, -10) -- node[below] {$t_1$} (axis cs:\Ftone+\Ftzero, -10);
			
			\draw[<->, thick] (axis cs:\Ftone+\Ftzero, -10) -- node[below] {$t_2$} (axis cs:\Ftone+\Fttwo+\Ftzero, -10);
			
			\draw[<->, thick] (axis cs:\Ftone+\Fttwo+\Ftzero, -10) -- node[below] {$t_3$} (axis cs:\Ftone+\Fttwo+\Ftthree+\Ftzero, -10);
			
			\draw[<->, thick] (axis cs:\Ftone+\Fttwo+\Ftthree+\Ftzero, -10) -- node[below] {$t_4$} (axis cs:\Ftone+\Fttwo+\Ftthree+\Ftzero+\Ftfour, -10);
			
			\addplot[only marks, mark=*, red] coordinates {(9.5, 15.206)} node[pin=135:{$S^i_{1}$}]{};
			
			\addplot[only marks, mark=*, red] coordinates {(12, 30.825)} node[pin=93:{$S^i_{2}$}]{};
			
		\end{axis}
	\end{tikzpicture}
	\vspace{-0.5em}
	\caption{Displacement vs. time graph for a drone travelling in a straight line with a start delay and end waiting time}
	\label{fig:generic DT diagram}
\end{figure}
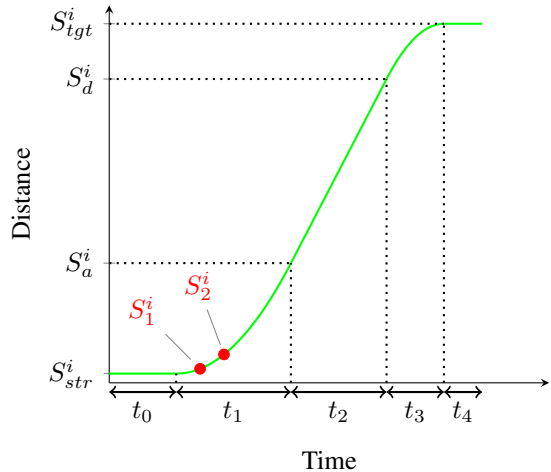

To address this challenge, the Time-efficient Prioritised Scheduling (TPS) method introduces a start-delay–based scheduling algorithm designed to eliminate collisions between drones while minimising the overall flocking time ($t_{\text{flock}}^n$). As previously discussed, each drone $i$ follows a straight-line trajectory subject to predefined physical constraints. However, following straight-line paths without coordination will inevitably lead to inter-drone collisions. Since the drone paths are fixed as straight lines, adjusting individual trajectories through calculated start time delays serves as an effective strategy to eliminate such collisions. The proposed movement mechanism is depicted in the velocity-time diagram in Fig. \ref{fig:generic VT diagram} and the corresponding displacement-time diagram is illustrated in Fig. \ref{fig:generic DT diagram}. In particular, the velocity–time diagram in Fig.~\ref{fig:generic VT diagram} incorporates a start delay of $t_0$, which shifts the original velocity–time profile shown in Fig.~\ref{fig:generic VT basic diagram} to the right by $t_0$. The displacement-time diagram shows the designated starting ($S^i_\text{str}$)  and target ($S^i_\text{tgt}$) positions. The drone will pass through various trajectory points, including the acceleration ending position ($S^i_a$) and the deceleration starting position ($S^i_d$) in the operational space, when reaching the target position. The red dots in Fig. \ref{fig:generic DT diagram} indicate some of these trajectory points.

As previously mentioned, acceleration time ($t_1$), maximum velocity time ($t_2$), and deceleration time ($t_3$) will be constant. The only variable affecting flocking time will be initial stationary time ($t_0$). 

The trajectory of any drone $i$ can be calculated using the equations from case \eqref{eq:path points}, where $t$ represents the elapsed time from the moment the first drone in the flock begins its motion.

\begin{equation}
	\begin{aligned}
		S^i(t) = 
		\begin{cases} 
			S^i_\text{str}, & \text{if } t \leq t_0^i  \\
			S^i_\text{str} + 0.5 \cdot a^i(t) \cdot (t-t^i_0)^2, & \text{if } t_0^i < t \leq t_0^i + t_1^i \\
			S^i_{a} + V^i(t)\cdot (t-(t^i_0+t^i_1)), & \text{if } t_0^i + t_1^i < t \leq t_0^i + t_1^i + t_2^i \\
			S^i_{d} + V^i(t)\cdot  (t -(t_0^i + t_1^i + t_2^i) \hspace{0.3cm} & \text{if } t_0^i + t_1^i + t_2^i < t \\ + 0.5 \cdot d^i(t) \cdot (t -(t_0^i + t_1^i + t_2^i))^2, & \hspace{0.2cm} \leq t_0^i + t_1^i + t_2^i + t_3^i \\
			S^i_\text{tgt}, & \text{if }  t_0^i + t_1^i + t_2^i + t_3^i < t
		\end{cases}
	\end{aligned}
	\label{eq:path points}
\end{equation}

The proposed solution should avoid any collisions between drones by introducing different starting delays for $k$ drones ($k < n$). Remaining $n-k$ drones will not have a starting delay ($t_0 = 0$). Let's assume that the $l-$th drone is the last to stop its movement since the first drone started its movement. Then the flocking time is given by \eqref{eq:total_time_flock}  

\begin{equation}
	t_{\text{flock}}^n = t_0^l + t_1^l + t_2^l + t_3^l
	\label{eq:total_time_flock}
\end{equation}

where, $t_0^l$ is the start delay of $l-$th drone.

Now that the key strategy for avoiding collisions has been finalised, two major questions remain for the algorithm to address. The first question is: if a potential collision between two drones exists, which drone should delay its movement? The second question is: what is the magnitude of the start delay that the selected drone must apply to its trajectory? Thus, based on the two questions outlined above, the original problem is decomposed into two sub-problems. As explained, when scheduling is the key strategy for resolving collisions, a hierarchy should be established between drones to determine the priority between any two drones. The first subproblem will propose a novel method to determine which drone can continue its original trajectory without introducing a starting delay. Once the hierarchy is determined, if there is a potential collision between two drones, the movement of the lower priority drone should be scheduled with a delay, ensuring collision-free travel. The second subproblem will propose a method to calculate individual delays for lower priority drones while attempting to keep the delays to a minimum.

\subsection{Drone Hierarchy Calculation}

When determining the hierarchy, the key consideration for giving priority is the higher possibility of any given drone colliding with another drone in the initial trajectory it takes. Since every drone travels in a straight line from start to target, individual paths are fixed for all drones. Thus, for a given drone, if one or more other drone paths are crossing its path with a distance less than the collision radius ($R_{\text{Col}}$), collisions can occur if the drone trajectories are not properly scheduled. Another related consideration is that once a drone reaches its target and becomes stationary, it can block other drones from travelling in their paths. This is because when a drone reaches its destination, a sphere with a radius $R_{\text{Col}}$ and a centre point with target coordinates acts as a fixed obstacle. To ensure none of the drones become fixed obstacles, they should not arrive at their respective destinations until all the other drones that pass within a distance of collision radius to their target are beyond this designated collision space. Similarly, a drone can act as a blocking drone at the starting position as well. In this case, a sphere with a radius $R_{\text{Col}}$ and a centre point with starting coordinates acts as a fixed obstacle. The solution in this scenario is to allow the blocking drone to move first, followed by the blocked drones. In summary, a blocked drone should receive priority over a blocking drone when potential collision may occur at the target position of the blocking drone. Conversely, a blocking drone should have higher priority over a blocked drone when the potential collision may occur at the staring position of the blocking drone.

\begin{figure}[h]
	\centering
	\includegraphics[width=0.45\textwidth]{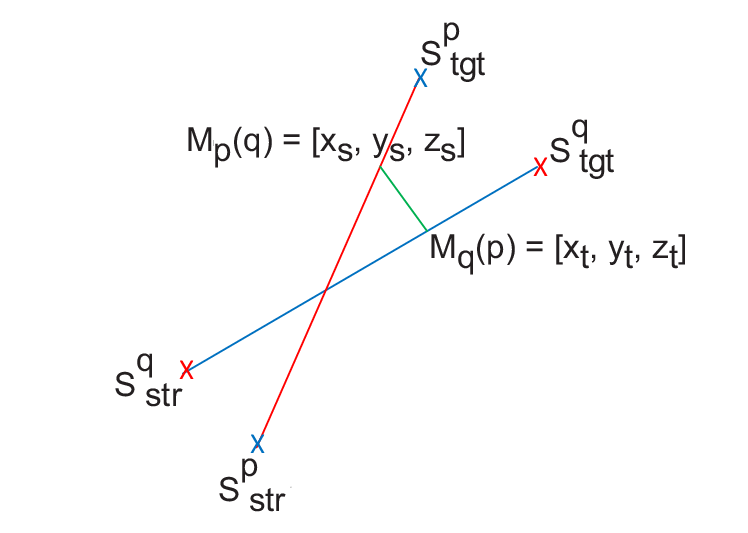}
	\caption{Example of the shortest distance between any two line segments.}
	\label{fig:two_lines_scenario}
\end{figure}

Let us assume that the paths of two drones $p$ and $q$ are presented
as two line segments according to Fig.~\ref{fig:two_lines_scenario}. $p$ and $q$ are index numbers given to any drone. When analysing any two drones, since $p \neq q$, the drone with a smaller index number is defined as $p$ and the other drone is defined as $q$. The shortest distance between two given line segments can occur between any two points in the line segment. According to Eberly \cite{Eberly_2018}, unless the paths of drones $p$ and $q$ are parallel, there will be a single line in the 3D space that provides the shortest distance between these two paths. For the drone $p$, the path position closest to the path of the drone $q$ is denoted as $M_p(q)$, and for the drone $q$, the path position is denoted as $M_q(p)$. Consider these path points as critical path points.

Eberly formulated a set of equations to calculate the path positions at which the shortest distance occurs between these two lines, which provides the coordinates of the path positions, $M_p(q)$ and $M_q(p)$. Once the path positions are known, the shortest distance between the two lines, that is, the distance between the positions $M_p(q)$ and $M_q(p)$, can be calculated.

The direction of each line is calculated using ~\eqref{eq:b1}
and \eqref{eq:b2}:
\begin{equation}
	\vec{b_1} = S^p_\text{tgt} - S^p_\text{str}
	\label{eq:b1}
\end{equation}
\begin{equation}
	\vec{b_2} = S^q_\text{tgt} - S^q_\text{str}
	\label{eq:b2}
\end{equation}
The parameters required for the calculation are given by ~\eqref{eq:a} and \eqref{eq:e}:
\begin{equation}
	a = \inner{\vec{b_1}}{\vec{b_1}}
	\label{eq:a}
\end{equation}
\begin{equation}
	b = \inner{\vec{b_1}}{\vec{b_2}}
	\label{eq:b}
\end{equation}
\begin{equation}
	c = \inner{\vec{b_2}}{\vec{b_2}}
	\label{eq:c}
\end{equation}
\begin{equation}
	d = \inner{\vec{b_1}}{\left( S^p_\text{str} - S^q_\text{str} \right)}
	\label{eq:d}
\end{equation}
\begin{equation}
	e = \inner{\vec{b_2}}{\left( S^p_\text{str} - S^q_\text{str} \right)}
	\label{eq:e}
\end{equation}
where $\inner{\cdot}{\cdot}$ denotes the inner product between two
vectors.

Assume that $s$ and $t$ are parameters within the range $[0, 1]$, which refers to a rescaled arc-length parameterisation of the trajectories (or curves) for drones $p$ and $q$, respectively. Then positions $M_p(q)$ and $M_q(p)$ can be defined using ~\eqref{eq:s_point} and \eqref{eq:t_point}:
\begin{eqnarray}
	\label{eq:s_point}
	M_p(q) & = & S^p_\text{str} + s \cdot \vec{b_1}  \\
	\label{eq:t_point}
	M_q(p) & = & S^q_\text{str} + t \cdot \vec{b_2} 
\end{eqnarray}
The parameters $s$ and $t$ that minimise the distance between positions on the paths are calculated as:
\begin{equation}
	s_{p,q} = \frac{b \cdot e - c \cdot d}{a \cdot c - b^2}
	\qquad 0 \leq s \leq 1
	\label{eq:s}
\end{equation}
\begin{equation}
	t_{p,q}  = \frac{a \cdot e - b \cdot d}{a \cdot c - b^2}
	\qquad 0 \leq t \leq 1
	\label{eq:t}
\end{equation}
If these parameters are outside the range $[0, 1]$, they are adjusted to be within this range by limiting them to the upper or the lower bound. To further explain assume that $s_{p,q} = 0.xyz$. This means the shortest distance between drone $p$ and drone $q$ occurs when drone $p$ has completed $xy.z\%$ of its journey. Also note that for any drone pair, $s_{p,q} = t_{q,p}$. Moreover, note that \eqref{eq:s} and \eqref{eq:t} can reach infinity when $ a\cdot c = b^2$. This corresponds to a situation where both lines are parallel, and such rare situations should be analysed separately. \par

Considering non-parallel paths, two positions on the paths that have the shortest distance between them are known; the Euclidean distance between those two positions will be the shortest possible distance between two drones. The minimum distance $\mu_{p,q}$ is then given by ~\eqref{eq:short_distance}:
\begin{equation}
	\mu_{p,q} = \norm{M_p(q) - M_q(p)}
	\label{eq:short_distance}
\end{equation}

Once the minimal distances between all drone pairs are calculated for $n$ drones, it can be stored in an $n$ by $n$ symmetric matrix $\mu$ \eqref{eq:distance_matrix}. Note that the elements in the table are symmetric across the diagonal and represent positive real numbers, as they correspond to distances between two positions. 

\begin{equation}
	\boldsymbol{\mu} =
	\begin{bmatrix}
		& \mu_{1,2} & \mu_{1,3} & \cdots & \mu_{1,n} \\
		\mu_{2,1} &  & \mu_{2,3} & \cdots & \mu_{2,n} \\
		\mu_{3,1} & \mu_{3,2} &  & \cdots & \mu_{3,n} \\
		\vdots & \vdots & \vdots & \ddots & \vdots \\
		\mu_{n,1} & \mu_{n,2} & \mu_{n,3} & \cdots & 
	\end{bmatrix}
	\label{eq:distance_matrix}
\end{equation}

Based on the $\mu_{p,q}$ values, the possibility of collision values $Pb_{p,q}$ are indicator variables telling whether a collision risk exists between drones $p$ and $q$. They are calculated using ~\eqref{eq:coll-prob} by comparing the shortest distance
between two drones $\mu_{p,q}$ against the collision radius $R_\text{col}$:
\begin{equation}
	Pb_{p,q} = 
	\begin{cases} 
		1, & \text{{if }} \mu_{p,q} \leq (SF \cdot R_\text{col})\\
		0, & \text{{if }} \mu_{p,q} > (SF \cdot R_\text{col}) \\
	\end{cases}
	\quad \text{for } p, q \in \{1, 2, \ldots, n\} \text{ and } p \neq q
	\label{eq:coll-prob}
\end{equation}
where $SF > 1$ is the safety factor (a given parameter). The safety factor ensures that marginal non-colliding instances are treated as potential collisions, providing a margin to compensate for small disturbances or measurement noise in drone positions.

Having a 1 in ~\eqref{eq:coll-prob} does not necessarily mean that the two corresponding drones will collide. The meaning of the 1 is that there exists a range of start-time differences between two drones that can cause a collision. However, if there is a 0 in the equation, it means that for the given starting and target positions of drone $p$ and $q$, two drones do not collide, irrespective of their start delays.

The example shown in Fig. \ref{fig:two_lines_scenario} illustrates one of eight possible relationship scenarios between two line segments. A comprehensive description of all eight scenarios is provided in Table \ref{tab:scenarios}. The terms \textit{hard constraint} and \textit{soft constraint} used in the table are defined as follows. A \textit{hard constraint} arises when two drones are on a potential collision course and a collision-free trajectory can only be guaranteed if one drone is explicitly assigned to move before the other. In contrast, a \textit{soft constraint} occurs when a potential collision can be avoided without enforcing a fixed priority between the two drones; that is, both ordering options remain feasible.

\begin{table}[H]
	\centering
	\caption{Possible configurations for two drones with respect to drone $p$}
	\vspace{1em}
	\begin{tabularx}{\textwidth}{|p{0.5cm}|p{3.2cm}|X|p{2.3cm}|}
		\hline
		\textbf{ID} & \textbf{Configurations} & \textbf{Description} & \textbf{Raw data}\\ \hline
		1 & \vspace*{\fill}\includegraphics[width=30mm]{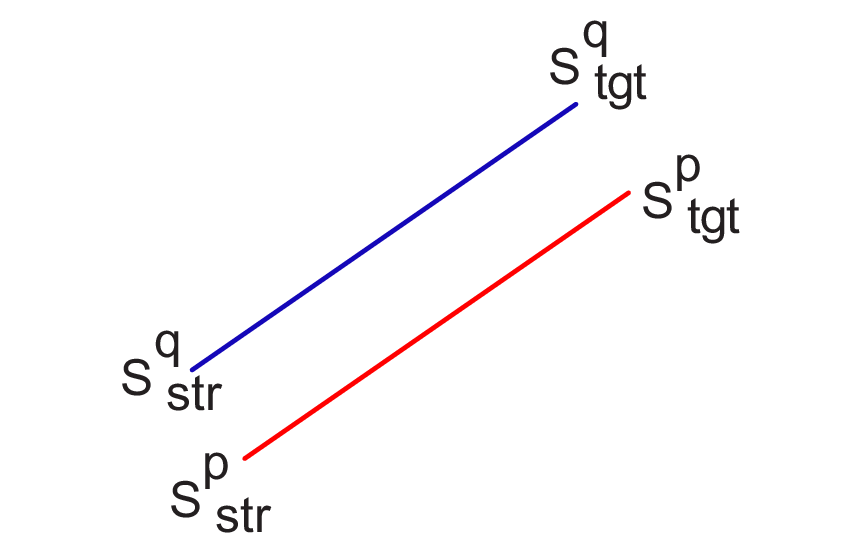}  
		& Two parallel drone paths with distance exceeding the collision threshold pose no collision risk. Parallel configuration is identified when the denominator of \eqref{eq:s} equals zero. & $\mu_{p,q} > SF \times R_\text{col}$ \newline $s,t \approx \infty $ \\ \hline
		
		2 & \vspace*{\fill}\includegraphics[width=30mm]{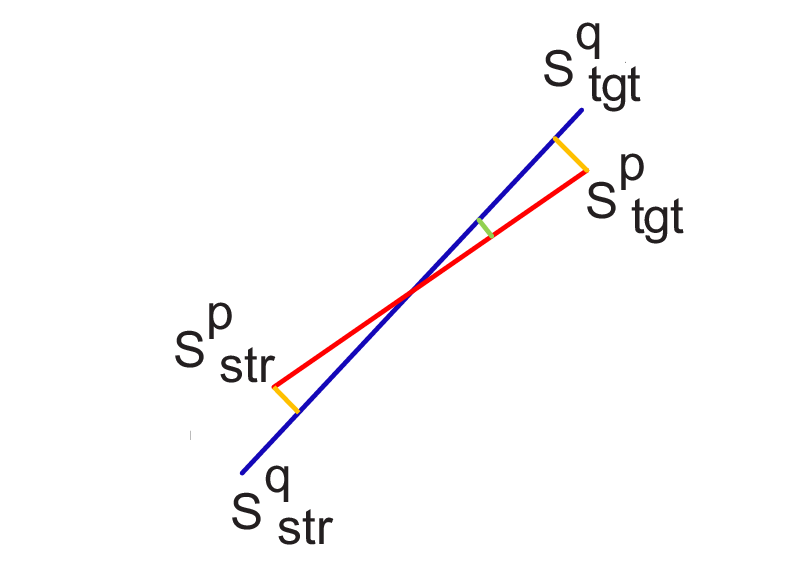}  
		& The whole path of drone $p$ is at a shorter distance than the collision threshold to the path of drone $q$. Scheduling cannot resolve this conflict.& $\mu_{p,q} < SF \times  R_\text{col}$ \newline at all times for drone $p$ \\ \hline
		
		3 & \vspace*{\fill}\includegraphics[width=30mm]{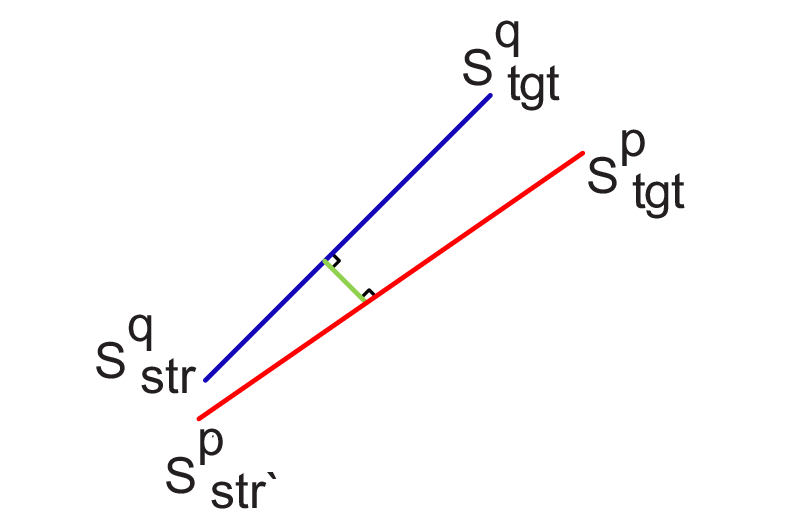} & When the shortest distance exceeds the collision threshold at all positions, this collision-free scenario requires no priority calculations. &$\mu_{p,q} > SF \times  R_\text{col}$ \newline $0 \leq s \leq 1 \newline 0 \leq t \leq 1 $
		\\ \hline
		
		4 & \vspace*{\fill}\includegraphics[width=30mm]{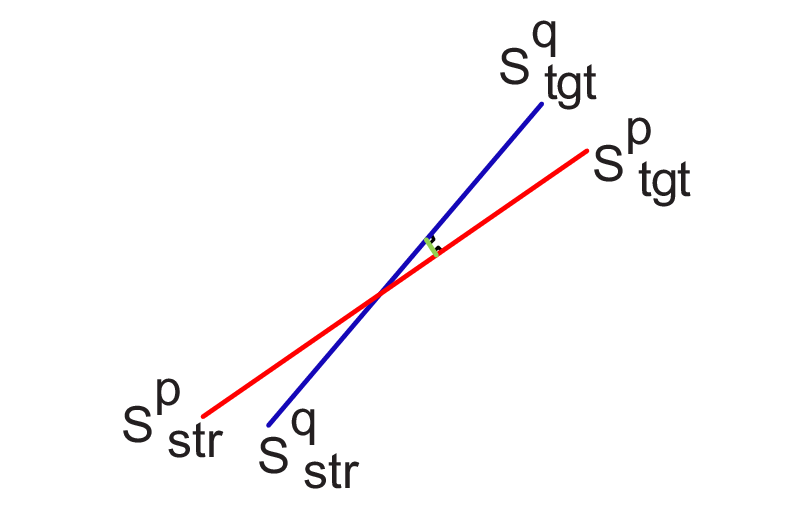} & When the shortest distance falls below the collision threshold at a middle position for drone $p$, while both staring position $S^p_\text{str}$ and target position $S^p_\text{tgt}$ exceed the threshold distance from drone $q$, the respective $s$ and $t$ values define a collision-range scenario. This creates a specific time delay range for drone $p$ causing collision, with delays outside this range remaining collision-free constituting a soft constraint. & $\mu_{p,q} <  SF \times R_\text{col}$ \newline $0 < s < 1 \newline 0 < t < 1 $ \\ \hline
		
		5 & \vspace*{\fill}\includegraphics[width=30mm]{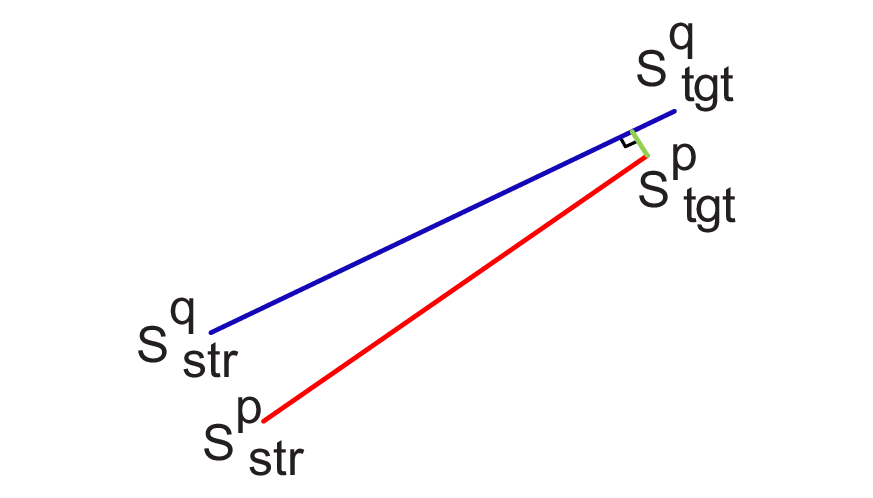} &  When the shortest distance falls below the collision threshold at drone $p$'s target position (a middle position for drone $q$), drone $p$ has value $1$ from ~\eqref{eq:s} while drone $q$ has value $t$ from ~\eqref{eq:t}, representing a collision limit scenario. Drone $p$ must arrive only after drone $q$ exits the collision radius of position $S^p_\text{tgt}$, constituting a hard constraint. &  $\mu_{p,q} <  SF \times R_\text{col}$ \newline $s = 1 \newline 0 < t < 1 $ \\ \hline  
		
		6 & \vspace*{\fill}\includegraphics[width=30mm]{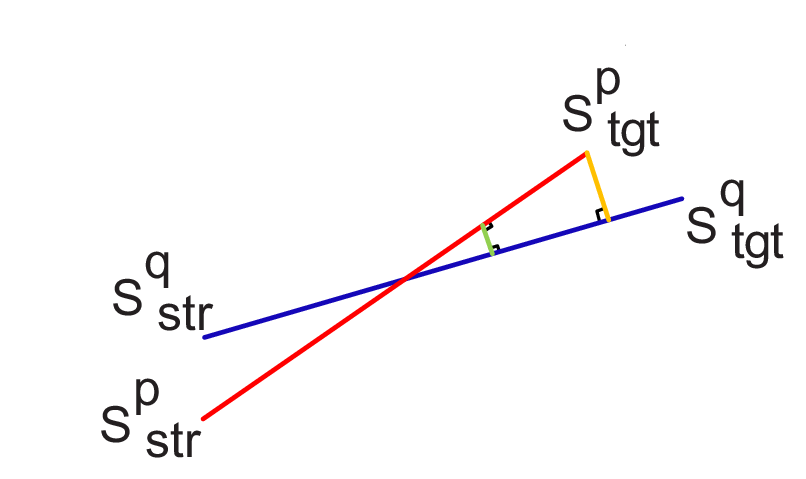} & When the shortest distance falls below the collision threshold at a middle position for both drones, with drone $p$'s target position $S^p_\text{tgt}$ also within the collision threshold of drone $q$'s path, drone $p$ must wait until drone $q$ exits the collision radius of position $S^p_\text{tgt}$, constituting a hard constraint like configuration 5. & $\mu_{p,q} <  SF \times R_\text{col}$ \newline $0 < s < 1 \newline 0 < t < 1 $  \\ \hline

		7 & \vspace*{\fill}\includegraphics[width=30mm]{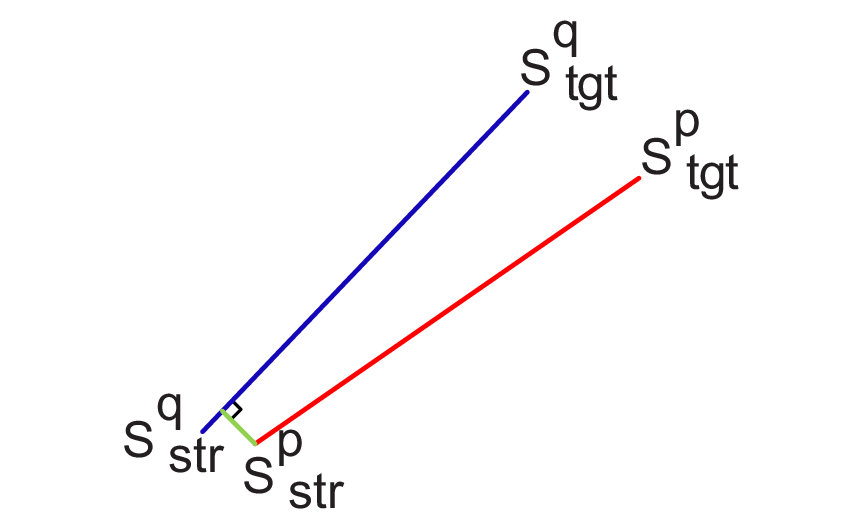} & When the shortest distance falls below the collision threshold at drone $p$'s starting position, drone $p$ must begin movement before drone $q$ enters its collision radius. This constitutes a hard constraint requiring drone $p$ to start its movement first. & $\mu_{p,q} <  SF \times R_\text{col}$ \newline $s = 0 \newline 0 < t < 1 $ \\ \hline
		
		8 & \vspace*{\fill}\includegraphics[width=30mm]{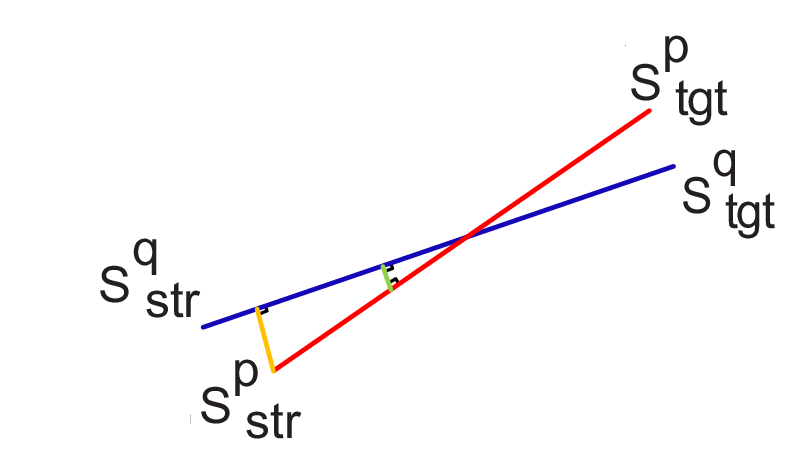} & When the shortest distance falls below the collision threshold at a middle position, with drone $p$'s starting position also within the collision threshold, drone $p$ must begin movement before drone $q$ enters its collision radius, constituting a hard constraint like scenario 7. & $\mu_{p,q} <  SF \times R_\text{col}$ \newline $0 < s < 1 \newline 0 < t < 1 $\\ \hline
		
	\end{tabularx}
	\label{tab:scenarios}
\end{table}

Using the results of ~\eqref{eq:s} and ~\eqref{eq:t}, an $n$ by $n$ matrix is generated to store the closest position relationships between each pair of drones. Unlike the distance matrix ($\mu$ matrix), the relationship matrix  ($R$ matrix) \eqref{eq:col_pos_matrix} is not symmetric, since the shortest distance between two drones may occur at different path positions for each drone. Hence, matrix $R$ is defined as follows:

\begin{equation}
	\boldsymbol{R} =
	\begin{bmatrix}
		& s_{1,2} & s_{1,3} & \cdots & s_{1,n} \\
		t_{1,2} &  & s_{2,3} & \cdots & s_{2,n} \\
		t_{1,3} & t_{2,3} &  & \cdots & s_{3,n} \\
		\vdots & \vdots & \vdots & \ddots & \vdots \\
		t_{1,n} & t_{2,n} & t_{3,n} & \cdots & 
	\end{bmatrix}
	\label{eq:col_pos_matrix}
\end{equation}

In the $R$ matrix, each data point (element) $R_{p,q}$ represents the position along drone $p$'s path that is closest to drone $q$. If $R_{p,q}$ equals zero, then from \eqref{eq:s_point}, $M_p(q)$ corresponds to $S^p_\text{str}$. Conversely, if it equals one, then using \eqref{eq:b1} and \eqref{eq:s_point}, $M_p(q)$ corresponds to $S^p_\text{tgt}$. When $R_{p,q}$ is a value between zero and one, $M_p(q)$ lies between the starting position and the target position. For a any given $R_{p,q}$ value, there exist a corresponding possibility of collision value $Pb_{p,q}$. Lets define a $S$ matrix to include the collision possibility between two drones using \eqref{eq:R_update},

\begin{equation}
	S_{p,q} = R_{p,q} \cdot Pb_{p,q} \quad \text{for all } p \neq q
	\label{eq:R_update}
\end{equation}

By updating the matrix using \eqref{eq:R_update}, effects of Configurations 1 and 3 in Table \ref{tab:scenarios} are excluded from calculations.

As shown in Table \ref{tab:scenarios}, there exist two types of constraints when assigning priorities. If it is a hard constraint, one drone must always have priority over the other drone. In contrast, if it is a soft constraint, priority is flexible as long as drones do not collide with each other. Given the above definitions, hard constraints under Configurations 6 and 8 exhibit numerical characteristics of soft constraints in matrix $S$. Therefore, a comprehensive collision matrix ($CL$) is defined using the $S$ matrix to address these edge cases using \eqref{eq:s-adjust},

\begin{equation}
	CL_{p,q} = 
	\begin{cases} 
		1, & \text{{if }} \mu_{p,q} \leq (SF \times R_\text{col}) \quad \mbox{at} \quad S^p_\text{tgt} \mbox{ or } S^q_\text{str}\\
		\lambda, & \text{{if }} \mu_{p,q} \leq (SF \times R_\text{col}) \quad \mbox{at} \quad S^p_\text{str}\\
		S_{p,q}, & \text{{otherwise }} 
	\end{cases}
	\label{eq:s-adjust}
\end{equation}
\noindent where \( \lambda \in (0, 1) \) is a real-valued constant.

In ~\eqref{eq:s-adjust}, each case corresponds to a specific edge condition:

\begin{itemize}
	\item \textbf{Case 1a:} When the target position of drone $p$, denoted as $S^p_\text{tgt}$, lies within the collision radius of drone $q$'s path, but $M_p(q) \neq S^p_\text{tgt}$, the value $CL_{p,q}$ is set to 1 to prioritize drone $q$. This adjustment is necessary because the original score $S_{p,q}$ indicates a non-blocking condition $(0 < S_{p,q} < 1)$. This modification corresponds to Configuration 6.
	
	\item \textbf{Case 1b:} When the starting position of drone $q$, $S^q_\text{str}$, lies within the collision radius of drone $p$'s path, and $M_p(q) \neq S^p_\text{tgt}$, the value $CL_{p,q}$ is set to 1 (where $(S_{p,q} = 1)$) to prioritise the departure of drone $q$ from $S^q_\text{str}$. This adjustment is necessary because the original score $S_{p,q}$ indicates a non-blocking condition $(0 < S_{p,q} < 1)$. This modification corresponds to drone $q$ configuration 8.

	\item \textbf{Case 2:} When the starting position of drone $p$, $S^p_\text{str}$, lies within the collision radius of drone $q$'s path, resulting in $M_p(q) = S^p_\text{str}$ and $(S_{p,q} = 0)$, the value $CL_{p,q}$ is adjusted to a small non-zero value ($\lambda$) to reflect a potential collision. Exact value of $\lambda$ is irrelevant as only requirement is to change the current zero value to a non-zero value to indicate a soft constraint. This adjustment is required because the original score $S_{p,q}$ suggests no collision when $(S_{p,q} = 0)$. This modification corresponds to drone $p$ in Configuration 8.
	
	\item \textbf{Case 3:} In all other scenarios, the original score $S_{p,q}$ is retained without modification.
\end{itemize}

After above adjustments, by examining the off-diagonal elements of matrix $CL$, specifically the data points $CL_{p,q}$ and $CL_{q,p}$, the potential collision relationship between drones $p$ and $q$ can be determined. There are three different characteristic groups in a $CL$ matrix, which are explained in the Definition \eqref{eq:CL_mat_cases},

\begin{equation}
	CL_{p,q} = 
	\begin{cases} 
		\text{No potential collision between drones} & \text{if } CL_{p,q} = 0 \\[0.5em]
		\text{Potential collision, no blocking (soft constraint)} & \text{if } 0 < CL_{p,q} < 1 \\[0.5em]            
		\shortstack[l]{Potential collision with blocking, drone $q$ \\ should have a higher priority than drone $p$ \\ (hard constraint)} & \text{if } CL_{p,q} = 1    
	\end{cases}
	\label{eq:CL_mat_cases}
\end{equation}

Warnakulasooriya et al. \cite{Warnakulasooriya_2025} identified that if only hard constraints are considered ($i.e., CL_{p,q} = 1$) and all other data points are treated as zero, $CL$ matrix can be interpreted as an adjacency matrix that stipulates the dependencies between drones. Consequently, calculating priority among drones can be formulated as a topological sorting problem. Standard topological sorting problem can be solved using either Kahn's algorithm \cite{Kahn_1962} or depth-first search (DFS) based method \cite{Tarjan_1971}. In the same study, authors employed a collision matrix containing only hard constraints as input to DFS-based method to determine whether a feasible delay-based scheduling solution exists for a given combination of start and target position configurations. They concluded that if a circular dependency is present, delay-based scheduling becomes infeasible. \par

\begin{figure}[H]
	\centering
	\includegraphics[scale=0.5]{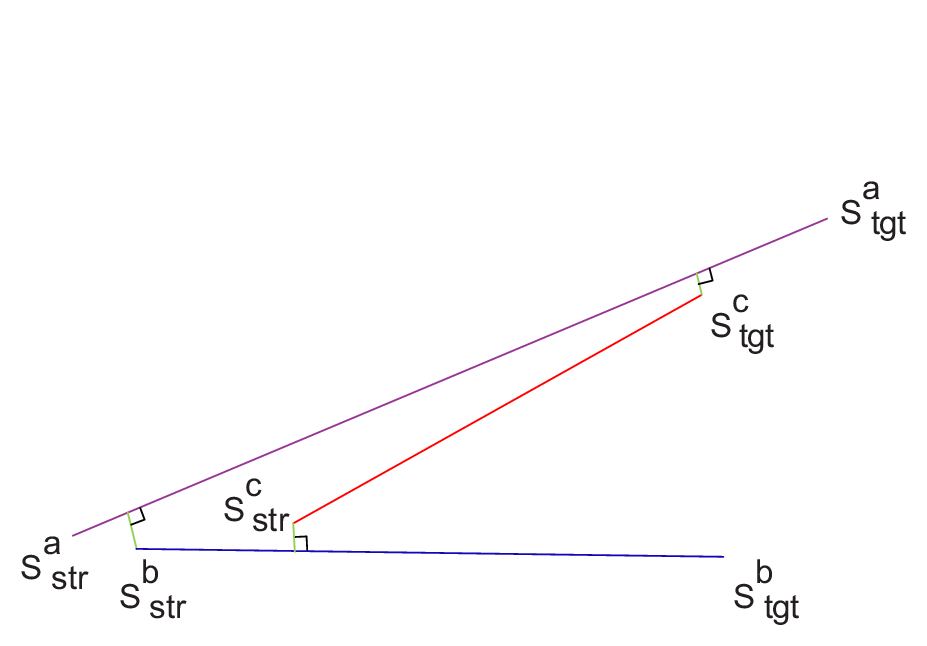}
	\caption{Example of a circular dependency}
	\label{fig:Scenario-2}
\end{figure}

If a cycle exists in a standard topological sorting problem, there will be no valid topological order, since circular dependencies cannot be resolved. A circular dependence may occur between three or more drones. In this research, circular dependencies could exist because all drones follow straight-line paths from the start position to the target position, and the only factor contributing to a circular dependency is the start and target position configuration combination. Fig.~\ref{fig:Scenario-2} showcases an example configuration of a circular dependence involving three drones.\par

In Fig.~\ref{fig:Scenario-2}, considering $S^b_\text{str}$ drone $b$ should depart before drone $a$ arrives near drone $b$’s starting position. Also, considering $S^c_\text{str}$ drone $c$ should depart before drone $b$ arrives near drone $c$’s starting position. However, considering $S^c_\text{tgt}$, drone $c$ should arrive at its destination only after drone $a$ passes its collision position. Since drone $a$ has a longer distance to travel, drone $a$ should depart before drone $c$ to avoid a collision. These three conditions cannot be satisfied at the same time resulting in a circular dependency. Thus, under the current scope, this condition is ruled out. DFS-based method is used to determine the existance of a cycle before the priority is determined. \par

Warnakulasooriya et al. \cite{Warnakulasooriya_2025} only investigated the feasibility of deploying a delay-based scheduling solution for a given drone configuration using standard DFS-based method. If no cycles exist, a DFS–based method can provide a priority that considers only the hard constraints. However, the existence of soft constraints makes the DFS-based method for determining the priority extremely inefficient, as it does not possess a mechanism to process soft constraints and their potential to cause collisions. Thus, an approach that incorporates both hard and soft constraints should be proposed to determine the priorities among drones. \par

Using the information in the $CL$ matrix, priority can be determined using the following steps considering both hard and soft constraints. As previously mentioned, there are two key characteristics when determining the hierarchy. The first characteristic is the number of other drone paths a given drone passes with less than the distance of the collision radius (total number of constraints). For any drone $k$, this number is provided by the number of non-zero values in row $k$ in the $CL$ matrix.  The other key characteristic is whether the given drone demands a lower priority than any other drone (number of hard constraints). This characteristic can be identified if there is a value $1$ in the row $k$. Until a priority is assigned to the drone that demands a higher priority, drone $k$ cannot be assigned a priority. Two phases are used in this calculation. Below are the steps of determining the priorities among $n$ drones:

\textbf{Phase I - Preprocessing:}\newline
\textbf{Step 1:} Create an Extended Collision (ECL) matrix of size $n \times (n+3)$ by augmenting the original Collision (CL) matrix with three additional columns:
\begin{itemize}
	\item Column $(n+1)$: \textbf{count column} - stores the count of soft constraints for each drone
	\item Column $(n+2)$: \textbf{max column} - stores the maximum value less than 1 in each row
	\item Column $(n+3)$: \textbf{blocks column} - stores the count of hard constraints for each drone
\end{itemize}

\noindent\textbf{Step 2:} For each drone $j$ (row $j$ in the matrix), calculate and store:
\begin{itemize}
	\item $ECL[j,n+1] = |\{k : 0 < CL(j,k) < 1\}|$ - the number of soft constraints (values between 0 and 1, exclusive)
	\item $ECL[j,n+2] = \max_{CL(j,k) < 1} CL(j,k)$ - the maximum value in the row that is less than 1
	\item $ECL[j,n+3] = |\{k : CL(j,k) = 1\}|$ - the number of hard constraints (values equal to 1)
\end{itemize}

\textbf{Phase II - Priority Calculation:}\newline
\textbf{Step 1:} The first iteration assigns priority value one, while each subsequent iteration assigns a priority value equal to the previous priority plus one. \newline
\textbf{Step 2:}: If all values in any row in $ECL$ matrix is equal to zero, that means that corresponding drone does not have any constraints. If any zero-constraint drones exist, select the first one (smallest index) and assign it to the priority vector $PV$. Skip the remaining steps for this iteration as the drone selection is complete. \newline
\textbf{Step 3:}: For each unprocessed drone, determine its eligibility for priority value allocation in the current round. A drone is excluded from priority value allocation in the current round if it has any hard constraints (i.e., $ECL[j,n+3] > 0$). \newline
\textbf{Step 4:} For eligible drones (those without hard constraints), use the precomputed soft constraint count from \textbf{count column} ($ECL[j,n+1]$). The drone with the highest count of soft constraints receives priority for the next priority assignment. This approach prioritises drones with the least flexibility, enabling them to select trajectories before more flexible drones. By planning first, highly constrained drones can secure feasible trajectories without being relegated to suboptimal conditions that require excessive delays due to higher-priority drone interference. \newline
\textbf{Step 5:} To resolve ties when multiple drones have identical counts of soft constraints, examine the precomputed maximum values in \textbf{max column} ($ECL[j,n+2]$) for the tied drones and select the drone with the smallest maximum value. For instance, if drones $x$, $y$, and $z$ have the same highest count $ECL[x,n+1] == ECL[y,n+1] == ECL[z,n+1]$, and $ECL[x,n+2] > ECL[z,n+2] > ECL[y,n+2]$, then drone $y$ is selected. The rationale is that smaller maximum values indicate potential collisions occurring further from the target position, and prioritising drones with more distant collision risks allows for better overall path optimisation. \newline
\textbf{Step 6:} The selected drone number is recorded in a priority vector. If the selected drone is $y$ and $m$ drones have already been assigned priorities, the priority vector $PV$ is updated as $PV(m+1) = y$.  \newline
\textbf{Step 7:} Eliminate the selected drone from future considerations by setting all values in its row to 0: $\forall i \in [1, n], ECL_{y,i} = 0$. This removes the influence of drone $y$ in subsequent iterations. \newline
\textbf{Step 8:} Update the matrix for incremental processing. Examine the column corresponding to the selected drone $y$. For any entry $ECL[j,y] = 1$ (hard constraint), update it to $ECL[j,y] = \gamma$ where $\gamma \in (0,1)$. Simultaneously update the precomputed columns:
\begin{itemize}
	\item Increase the soft constraint count: $ECL[j,n+1] = ECL[j,n+1] + 1$
	\item Decrease the hard constraint count: $ECL[j,n+3] = ECL[j,n+3] - 1$
\end{itemize}
This adjustment ensures that if drone $y$ previously acted as a blocking drone for drone $j$, effect of this blocking is reversed by considering drone $y$ as a soft constraint and not a hard constraint for drone $j$ going forward. \newline
\textbf{Step 9:} Repeat Steps 3-8 for all $n$ drones until the complete priorites are established for all drones.

The pseudocode for obtaining the priorities for any given set of drones is given by Algorithm \ref{alg:optimized_priority_matrix}. Consider any data point in Matrix $CL$ is denoted by $CL(i,j)$.


\begin{algorithm}
	\caption{Priority Calculation}
	\begin{algorithmic}[1]
		
		\State \textbf{Preprocessing:}
		\State Create $ECL[n \times (n+3)] = [CL \mid \text{count\_column} \mid \text{max\_column} \mid \text{blocks\_column}]$
		\For{$j = 1$ to $n$}
		\State $ECL[j,n+1] = |\{k : 0 < CL(j,k) < 1\}|$ \Comment{soft constraint count}
		\State $ECL[j,n+2] = \max_{CL(j,k) < 1} CL(j,k)$ \Comment{maximum value in row less than 1}
		\State $ECL[j,n+3] = |\{k : CL(j,k) = 1\}|$ \Comment{hard constraint count}
		\EndFor
		
		\State \textbf{Priority Calculation:}
		\State Initialize $PV = []$, $processed = [\text{false}]^n$
		\For{round $i = 1$ to $n$}	
		\If{$\exists j : ECL[j,n + 1] = 0 \text{ and } ECL[j,n + 3] = 0 \text{ and } processed[j] = \text{false}$}
		\State $selected = \min\{j : ECL[j,n + 1] = 0 \text{ and } ECL[j,n + 3] = 0 \text{ and } processed[j] = \text{false}\}$
		\State $PV[i] = selected$, $processed[selected] = \text{true}$, 
		\State \textbf{continue}
		\Comment{assign priority to drones without constraints}
		\EndIf

		\For{$j = 1$ to $n$ where $processed[j] = \text{false}$}
		\State $count[j] = 0 \text{ if }(ECL[j,n+3] > 0), \text{ otherwise } ECL[j,n+1]$
		\Comment{O(1) lookup}
		\EndFor
		\State $maxCount = \max(count)$
		\State $candidates = \{j : count[j] = maxCount \text{ and } processed[j] = \text{false}\}$
		\If{$|candidates| > 1$}
		\State $minValue = \min\{ECL[c,n+2] : c \in candidates\}$
		\State $selected = \{c \in candidates : ECL[c,n+2] = minValue\}$
		\Comment{O(1) tie-break}
		\Else
		\State $selected = candidates[1]$
		\EndIf
		
		\State $PV[i] = selected$, $processed[selected] = \text{true}$
		\State $ECL[selected,:] = 0$ \Comment{eliminate the effect of selected drone}
		
		\For{$j$ where $ECL[j,selected] = 1$} \Comment{incremental updates}
		\State $ECL[j,selected] = \gamma \in (0,1)$
		\State Update $ECL[j,n+1] = ECL[j,n+1] + 1$; \Comment{increase soft constraint count by 1}
		\State Update $ECL[j,n+3] = ECL[j,n+3] - 1$; \Comment{decrease hard constraint count by 1}
		\EndFor
		\EndFor
		\State \Return $PV$
		
	\end{algorithmic}
	\label{alg:optimized_priority_matrix}
\end{algorithm}

\subsection{Drone Collision Scenarios and Delay Calculation}

Once the priorities are established, the individual delay for each of the $n$ drones can be calculated. Similar to the hierarchy determination, the delay computation is performed pairwise. For any two drones, one will always have a higher priority and the other a lower priority. The proposed method calculates the necessary delay for the lower priority drone relative to all higher priority drones that are on a potential collision course. The lower priority drone then selects the minimum delay required to ensure a collision-free trajectory with all such higher prioritised drones.

Considering the relationship between the higher priority and lower priority drones, three different scenarios of delay calculations can be identified.

\subsubsection{Collision-free scenario}

\begin{figure}[H]
	\centering
	\includegraphics[width=0.7\textwidth]{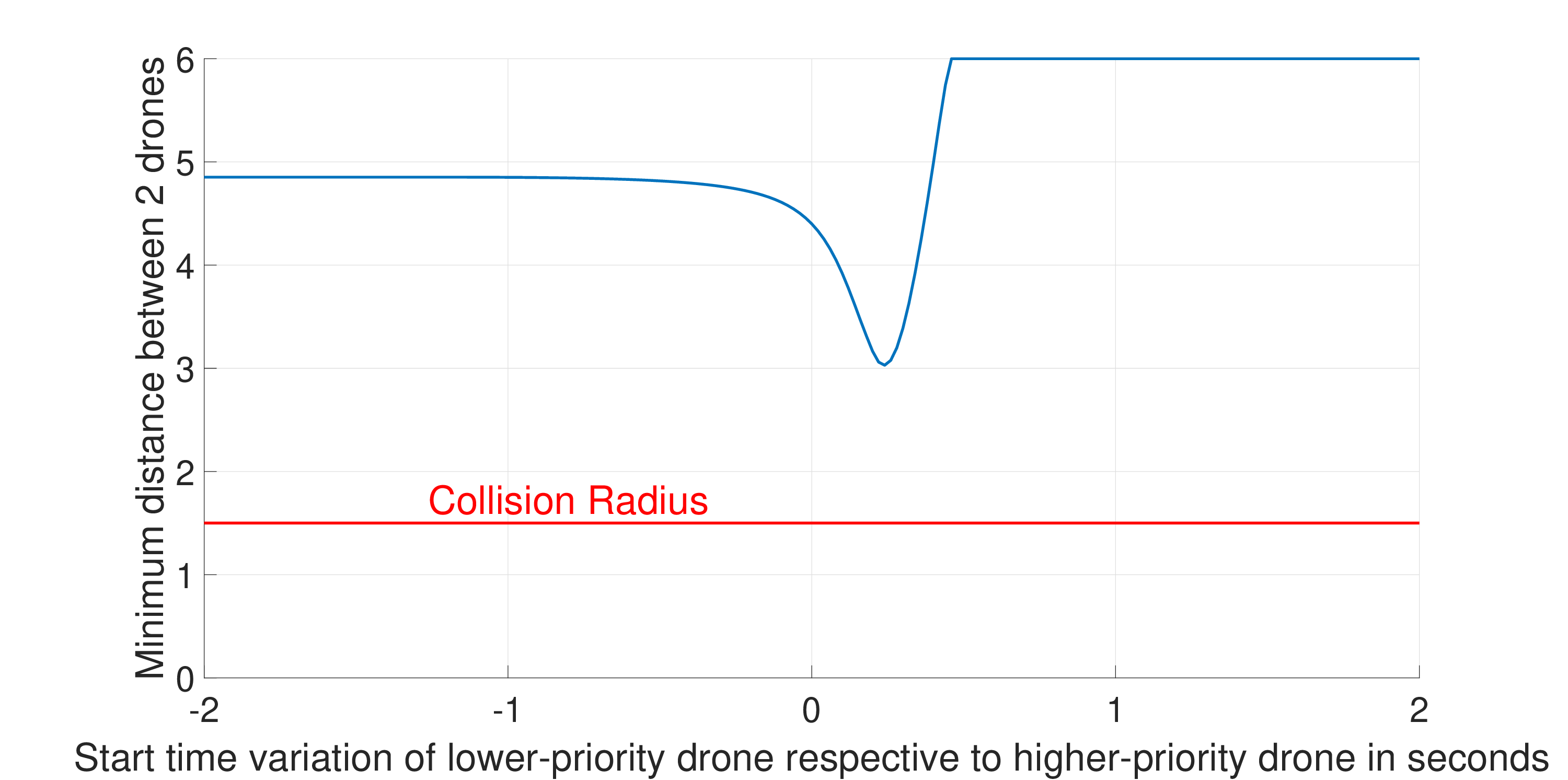}
	\caption{Minimum distance between higher priority drone with fixed start time vs lower priority drone with variable starting time under \textbf{collision-free scenario}}
	\label{fig:free_col}
\end{figure}

In this scenario, the two drones under consideration will never come closer than the collision radius ($R_\text{col}$). This situation corresponds to Configuration 1 or 3 in Table \ref{tab:scenarios}. Therefore, when calculating the delay for the lower priority drone, the effects of the higher priority drone can be disregarded. Fig. \ref{fig:free_col} illustrates the minimum distance between a higher priority drone and a lower priority drone. The starting time for the higher priority drone is fixed, while the starting time of the lower priority drone varies from $-1$ second to $1$ second relative to the fixed drone. The minimum distance between the two drones does not fall below the collision radius at any given time. Consequently, the lower priority drone's starting time is unaffected by the higher priority drone.

\subsubsection{Collision limit scenario}

\begin{figure}[H]
	\centering
	\includegraphics[width=0.7\textwidth]{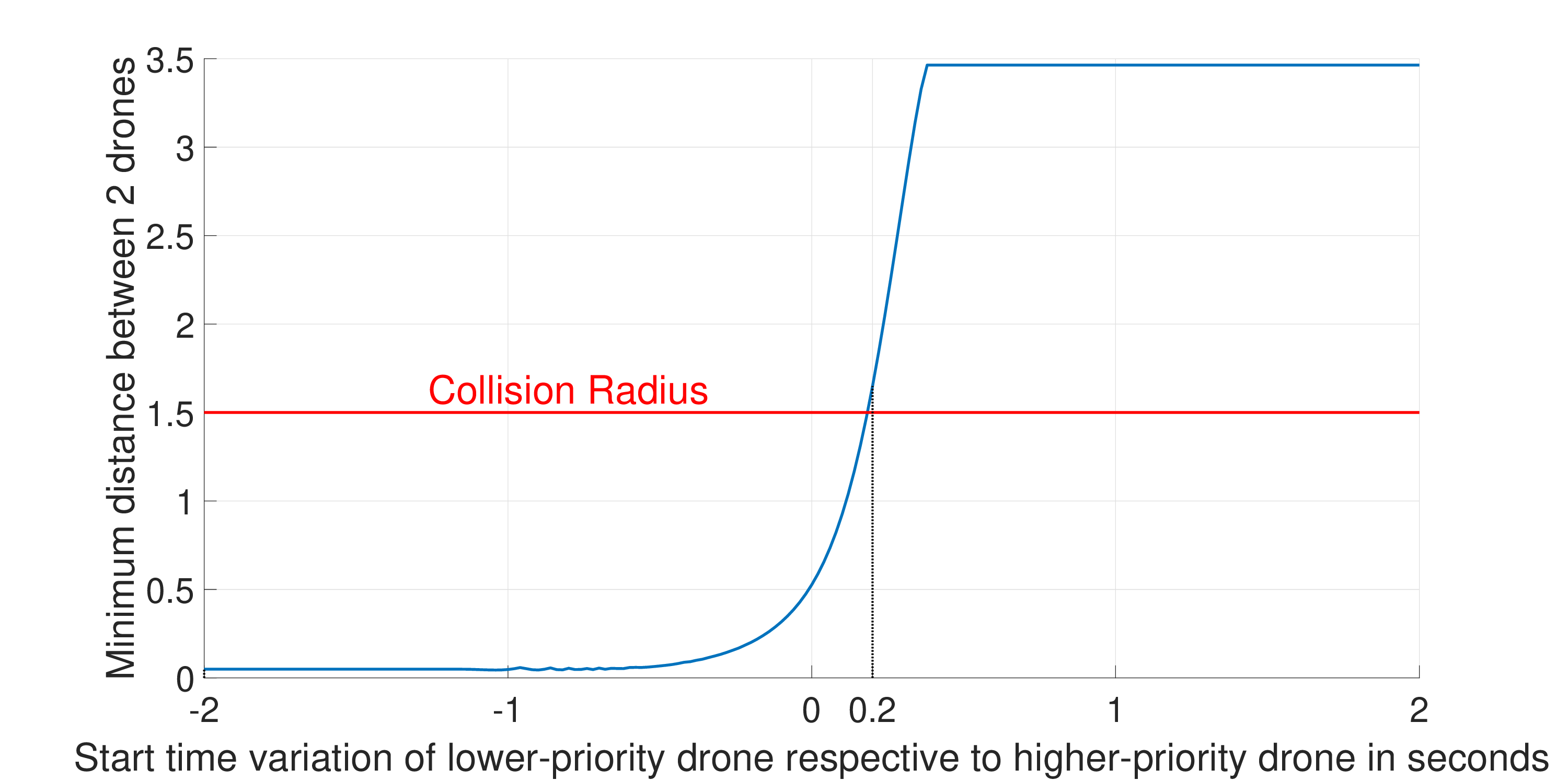}
	\caption{Minimum distance between higher priority drone with fixed start time vs lower priority drone with variable starting time under \textbf{collision limit scenario}}
	\label{fig:limit_col}
\end{figure}

A collision limit scenario occurs when the lower priority drone should allow the higher priority drone to move first in the potential collision zone. This scenario corresponds to configurations 5, 6, 7 or 8 in Table \ref{tab:scenarios}. In the $CL$ matrix, any data point $CL_{p,q}$ is equal to a value $1$, the row number indicates the lower priority drone, and the column number indicates the higher priority drone. To resolve this conflict, the departure of the lower priority drone must be delayed by a certain time threshold. According to Fig. \ref{fig:limit_col}, the lower priority drone must have a specific delay relative to the higher priority drone to ensure a collision-free journey. Any delay longer than the marginal delay is acceptable, but the minimum delay required for collision-free travel is the marginal delay.

\subsubsection{Collision range scenario}

\begin{figure}[H]
	\centering
	\includegraphics[width=0.7\textwidth]{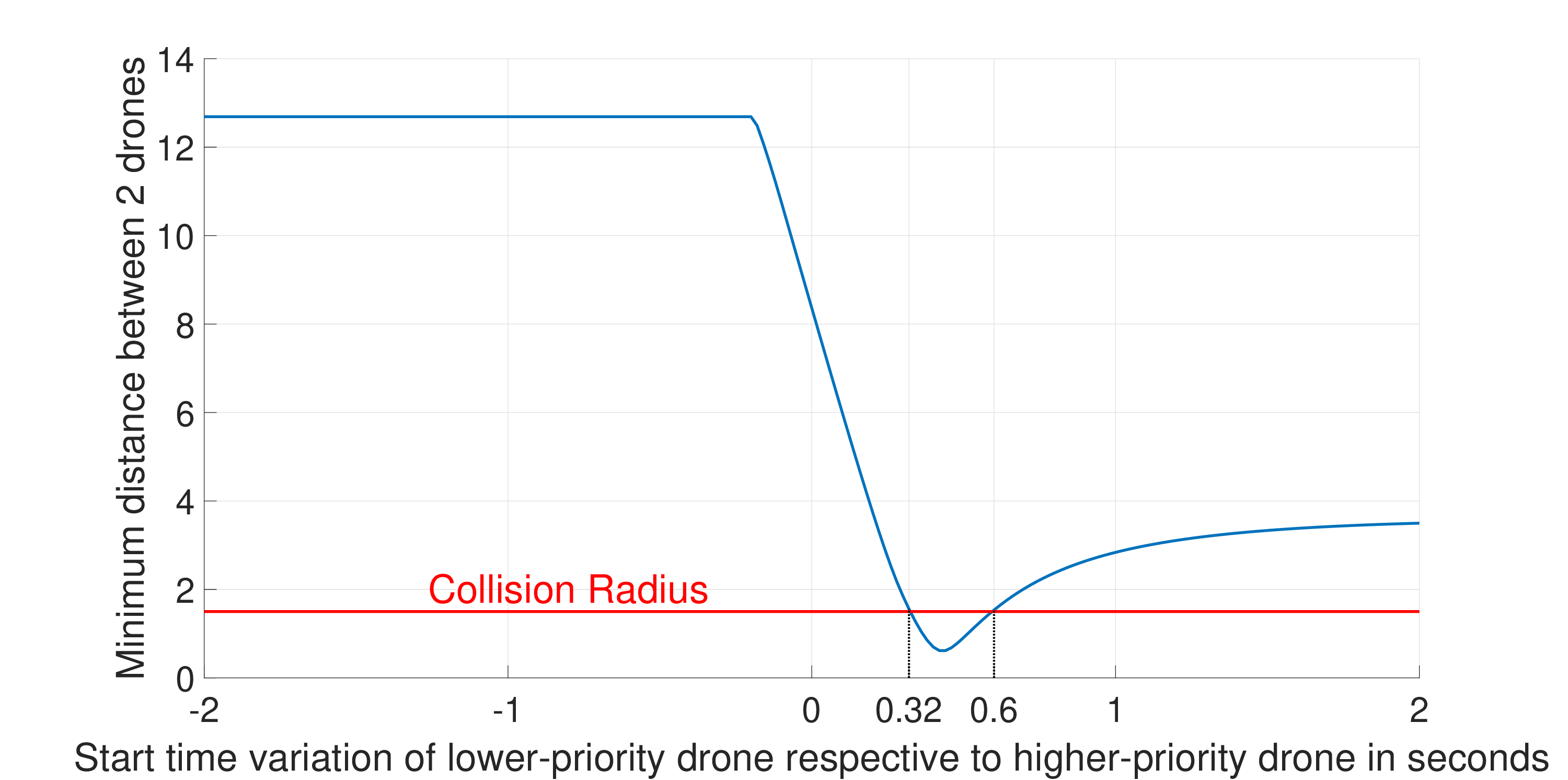}
	\caption{Minimum distance between higher priority drone with fixed start time vs lower priority drone with variable starting time under \textbf{collision range scenario}}
	\label{fig:range_col}
\end{figure}

A collision range scenario happens when there exist a specific range of time delays for the lower priority drone that could cause a collision. As long as the delay for the lower priority drone is outside this range, it will not cause a collision, and this corresponds to Configuration 4 in Table \ref{tab:scenarios}. As illustrated in Fig.~\ref{fig:range_col}, only a specific time delay range will cause a collision and a lower priority drone should avoid that range.

\subsubsection*{Drone delay calculation} 
\vspace{0.5em}
\noindent
Before presenting the delay calculation method, two types of delays must be defined. Absolute starting delay refers to the delay that any lower-priority drone may have relative to the highest-priority drone, and it cannot take a negative value. Relative starting delay denotes the delay that a lower-priority drone may have with respect to any given higher-priority drone. This value may be negative, provided that the absolute starting delay of the lower-priority drone remains non-negative.

\par
As collisions cannot occur if two drones have a collision-free relationship, a collision-limit scenario is considered first. Assume a higher priority drone $p$ and a lower priority drone $q$ exist in such a scenario. Using \eqref{eq:path points}, the distance between drones $D_{p,q}$ can be expressed as follows:
  
\begin{equation}
	D_{p,q}(t) = \|S^p(t)-S^q(t)\| 
	\label{eq:Min_Distance_Cal}
\end{equation}

Assume that $t^p_0 = 0$, and equation \eqref{eq:Min_Distance_Cal} must be solved subject to the constraint $\min(D_{p,q}) > SF \times R_\text{col}$ while minimising the starting delay $t^q_0$ for drone $q$. Let $\bar{t}^q_0$ denote the minimum upper-bound starting delay for drone $q$ that guarantees collision-free travel. However, obtaining the analytical solution for $\bar{t}^q_0$ is computationally intensive. Therefore, a numerical approach is proposed to determine $\bar{t}^q_0$.

\begin{figure}[h]
	\centering
	\includegraphics[width=0.55\textwidth]{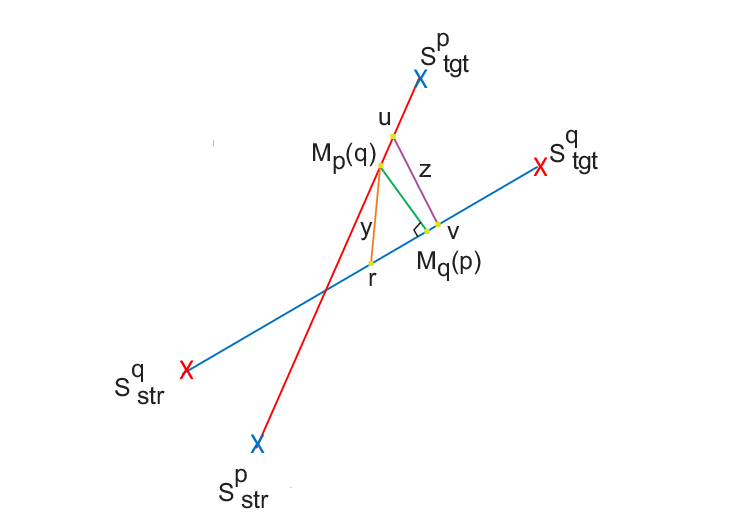}
	\caption{Intutive solution.}
	\label{fig:intutive}
\end{figure}

When calculating individual starting delays for a lower-priority drone, the critical travel time for both drones should be determined first. The critical travel time ($t_{cr}^i$) is the time taken for both drones to travel to the critical path points $M_p(q)$ or $M_q(p)$, as shown in Fig.~\ref{fig:intutive}. To solve for drone $p$, substitute the distance $M_p(q)$ for $S^p(t)$ in \eqref{eq:path points} to calculate the critical travel time $t_{cr}^p$. Similarly, for drone $q$, the critical travel time $t_{cr}^q$ can be calculated. The critical time difference for drone $q$ with respect to drone $p$, denoted as $T_{cr}^{q-p}$, is given by \eqref{eq:Crit_Time_Diff}. As $T_{cr}^{q-p}$ is a relative starting delay, its value may even be negative. A positive value of $T_{cr}^{q-p}$ indicates that drone $p$ requires a longer travel time to reach its critical path points compared to drone $q$, whereas a negative value indicates that drone $q$ requires a longer travel time.

\begin{equation}
	T_{cr}^{q-p} = t_{cr}^p - t_{cr}^q
	\label{eq:Crit_Time_Diff}
\end{equation}

An intuitive method of calculating the required starting time delay ($t^{q-p}_0$) for the lower-priority drone $q$ is as follows. Fig.~\ref{fig:intutive} provides an elaborated extension of Fig.~\ref{fig:two_lines_scenario}. Assume that the higher-priority drone $p$ is at $M_p(q)$, which denotes the closest point on drone $p$’s path to the path of drone $q$. First, calculate the travel time for drone $p$ to arrive at position $M_p(q)$. Next, identify a point $r$ on drone $q$’s path that lies before $M_q(p)$ and is at a distance greater than the collision threshold. In this scenario, the distance between $M_p(q)$ and $r$ is denoted by $y$, where $y > SF \times R_\text{col}$. Then, compute the travel time for drone $q$ to arrive at point $r$. The difference between the travel times of drone $p$ and drone $q$ can be taken as the starting time delay ($t^{q-p}_0$) for drone $q$. The reasoning behind this calculation is that if the separation between the two drones, when drone $p$ is at $M_p(q)$, exceeds the collision threshold, then every other point along their respective paths will also remain collision-free.

However, contrary to intuition, this approach is not always successful due to differences in velocity profiles. Two drones may be farther apart than the collision threshold at a critical point such as $M_p(q)$. Yet, if their velocities differ, other trajectory points may still result in collisions. Referring to Fig.~\ref{fig:intutive}, consider the case where, at a given instant, drone $p$ is at point $M_p(q)$ and drone $q$ is at point $r$, with the distance between them exceeding the collision threshold. From this instant, drones $p$ and $q$ travel for the same duration to arrive at points $u$ and $v$, respectively. The distance between these two points is $z$, where $z < SF \times R_\text{col}$, which could cause a collision. In summary, even if no collision occurs at the critical point $M_p(q)$, differences in velocity profiles can lead to collisions at other points along the trajectories. Therefore, to guarantee a collision-free path for the entire journey, the proposed binary search method becomes essential.

\par

For drone $q$, the following method is used to achieve the minimum upper-bound starting delay $\bar{t}^{q-p}_0$. If drone $q$ initiates its movement at the critical time difference ($T_{cr}^{q-p}$) relative to drone $p$, it will result in achieving the minimum distance $\mu_{p,q}$ between the two drones. The lower-priority drone $q$ can then be delayed incrementally by a predefined step size parameter ($ DT >0$) until a collision-free trajectory is guaranteed between the higher-priority drone $p$ and the lower-priority drone $q$. However, determining the minimum upper-bound starting delay $\bar{t}^{q-p}_0$ through iterative incrementing is highly inefficient. Therefore, a binary search approach is proposed to efficiently calculate $\bar{t}^{q-p}_0$ within an appropriate maximum allowable delay bound ($T^\text{max}_\text{D}$). The bound $T^\text{max}_\text{D}$ is calculated by considering the percentage of the journey completed by drone $q$ by the time it reaches position $M_q(p)$, with the maximum possible value of $T^\text{max}_\text{D}$ being $t_{cr}^q$.

\begin{equation}
	\bar{t}^{q-p}_0 = \text{BinarySearch}\left(T_{cr}^{q-p}, T_{cr}^{q-p} + T^\text{max}_\text{D}, \text{CollisionFree}\right)
	\label{eq:crit_time_diff_upper}
\end{equation}

Equation \eqref{eq:crit_time_diff_upper} defines the minimum upper-bound starting delay $\bar{t}^{q-p}_0$ for the lower-priority drone $q$ using a binary search procedure. The binary search is applied over an interval that begins at the critical time difference $T^{cr}_{q-p}$ and extends to $T^{cr}_{q-p} + T^\text{max}_\text{D}$, where $T^\text{max}_\text{D}$ is the maximum allowable delay bound. The search criterion is the function $\text{CollisionFree}$, which evaluates whether a given starting delay results in a collision-free trajectory between drones $p$ and $q$. In essence, the equation states that $\bar{t}^{q-p}_0$ is the smallest delay for drone $q$, found efficiently via binary search, that guarantees collision-free travel relative to drone $p$. The binary search avoids inefficient incremental checking by systematically halving the search space until the minimum safe delay is identified. \par

A similar approach can be applied when calculating marginal delays under a collision-range scenario. As there exists a time-delay range that can cause a collision, both an upper-bound and a lower-bound starting delay must be considered for drone $q$ in such a scenario. The upper-bound starting delay can be calculated using the same approach as in \eqref{eq:crit_time_diff_upper}. The lower-bound starting delay, denoted as $\underline{t}^{q-p}_0$, is obtained by modifying the search direction as described in \eqref{eq:crit_time_diff_lower}. It should be noted that when determining the lower-bound starting delay, the equation incorporates a minimum allowable delay bound of $-T^\text{max}_\text{D}$.

\begin{equation}
	\underline{t}^{q-p}_0 = \text{BinarySearch}\left(T_{cr}^{q-p}, T_{cr}^{q-p} - T^\text{max}_\text{D}, \text{CollisionFree}\right)
	\label{eq:crit_time_diff_lower}
\end{equation}

Equations \eqref{eq:crit_time_diff_upper} and \eqref{eq:crit_time_diff_lower} calculate the marginal starting delay bounds for a lower-priority drone against a single higher-priority drone. As these bounds are relative, they must be converted into absolute delays. This is achieved by adding the final delay of the corresponding higher-priority drone to the relative delay of the lower-priority drone. In this way, the absolute starting delay for the lower-priority drone ensures collision-free travel with respect to all higher-priority drones.

\begin{equation}
	t^q_0 = t^{q-p}_0 + t^p_0
	\label{eq:absolute_delay}
\end{equation}

\begin{figure}[H]
	\centering
	\includegraphics[width=0.8\textwidth]{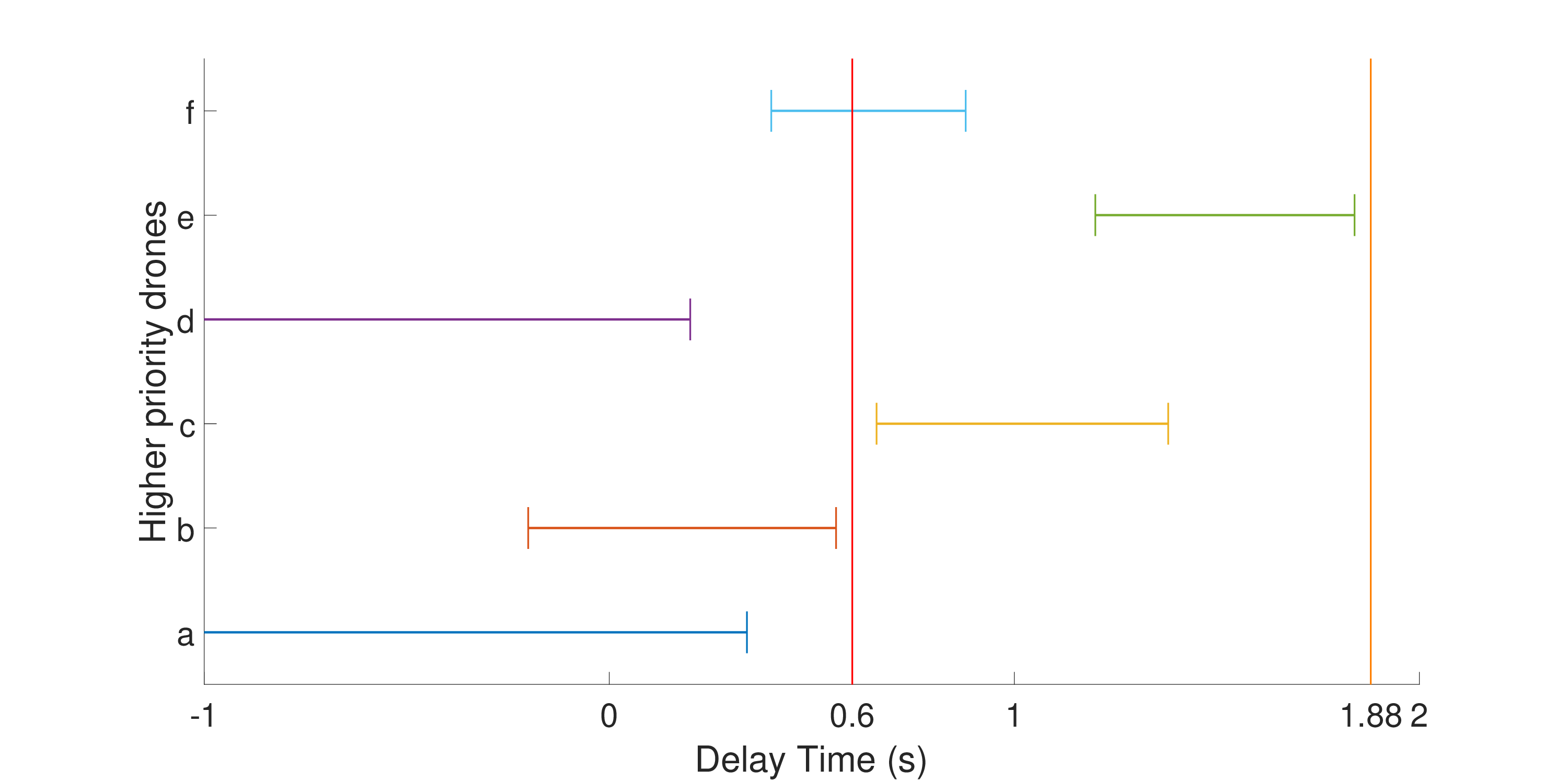}
	\caption{Delay calculation visual graph}
	\label{fig:delay_graph}
\end{figure}

Additionally, if there are multiple higher-priority drones that could cause a collision, the final starting delay for the lower-priority drone must guarantee collision-free travel with all of them. Fig.~\ref{fig:delay_graph} visually illustrates the absolute collision times for six higher-priority drones against a lower-priority drone. Drones $a$ and $d$ are in a collision-limit relationship, while the other four drones have a collision-range relationship with the given lower-priority drone. The graph shows the time delays that would cause a collision for a lower-priority drone with six higher-priority drones. Considering all six higher-priority drones, the orange line represents the minimum delay that avoids a collision. However, if drone $f$ did not exist, the minimum delay could be significantly reduced, as indicated by the red vertical line.

Let us assume that for a lower-priority drone $q$, there are $x$ higher-priority drones that can potentially cause a collision. For any given higher-priority drone $p$, the starting delay $t^{q}_0$ must not lie within the delay range [$\underline{t}^{q-p}_0$, $\bar{t}^{q-p}_0$]. Consequently, the minimum starting delay for drone $q$, denoted as $t^{q-\text{final}}_0$, is defined as the smallest non-negative starting delay such that the chosen delay is not contained within the union of all delay ranges corresponding to the higher-priority drones that share a collision trajectory with drone $q$.

\begin{equation}
	t^{q-\text{final}}_0= \min\left\{ t \geq 0 : 
	t \notin \bigcup_{i=1}^{x} \left[ \underline{t}^{q-i}_0 + t^i_0 \; , \; \bar{t}^{q-i}_0 + t^i_0 \right] \right\}
	\label{eq:lowest_outside_ranges}
\end{equation}

In summary, given below are the steps to follow when calculating the
time delay for all $n$ drones:
\begin{itemize}
	\item \textbf{Step 1:} Initialise the algorithm by setting the highest-priority drone’s delay time to zero and adding it to the planned drones list.
	\item \textbf{Step 2:} For each subsequent drone in priority order:
	\begin{itemize}
		\item \textbf{Collision identification:} Identify all previously planned drones within the collision safety-factor range ($\mu_{j,x} \le SF \cdot R_{\text{col}}$).
		\item \textbf{Range computation:} For each potentially conflicting drone pair, compute critical collision times and determine forbidden delay-time ranges using binary search (upper bound via \eqref{eq:crit_time_diff_upper}, lower bound via \eqref{eq:crit_time_diff_lower}).
	\end{itemize}
	\item \textbf{Step 3:} Process the forbidden delay ranges by:
	\begin{itemize}
		\item \textbf{Sort:} Order all forbidden ranges in ascending order by their lower bounds.
		\item \textbf{Select delay:} Find the minimum feasible non-negative delay time that avoids all forbidden ranges by checking for gaps between consecutive ranges; equivalently, use \eqref{eq:lowest_outside_ranges}.
	\end{itemize}
	\item \textbf{Step 4:} Assign the calculated minimum delay to the current drone and add it to the planned drones list.
	\item \textbf{Step 5:} Repeat Steps 2--4 for all remaining drones in descending priority order until all $n$ drones are assigned.
\end{itemize}

The pseudocode for obtaining the time delay for any given set of drones is given by Algorithm \ref{alg:timing_delay_cal}.

\begin{algorithm}
	\caption{Delay-Time Calculation}
	\begin{algorithmic}[1]
		\State Initialize $timeAdded = []$
		\State $y = PV[1]$
		\Comment{PV is the calculated priority using Algorithm \ref{alg:optimized_priority_matrix}}
		\State $timeAdded[y] = 0$
		\For{$i = 2$ to $n$}
			\State $x = PV[i]$
			\State Create $avoidRange[2 \times (i-1)] = []$
			 	\For{$j = PV[1:i-1]$}
			 		\If {$\mu_{j,x} \leq (SF \cdot R_\text{col})$}
			 			\State Calculate $t_{cr}^j$ using \eqref{eq:path points} with $S^j(t) = M_j(x)$  
						\State Calculate $t_{cr}^x$ using \eqref{eq:path points} with $S^x(t) = M_x(j)$
						\State $T_{cr}^{x-j} = t_{cr}^j - t_{cr}^x$ \Comment{Using \eqref{eq:Crit_Time_Diff}}
						\State $\bar{t}^{x-j}_0 = \text{BinarySearch}\left(T_{cr}^{x-j}, T_{cr}^{x-j} + T^\text{max}_\text{D}, \text{CollisionFree}\right) $
						\Comment{Using \eqref{eq:crit_time_diff_upper}}
						\If {$CL[x,j] < 1$}
							\Comment{if $x$ and $j$ are in a collision range scenario}
							\State $\underline{t}^{x-j}_0 = \text{BinarySearch}\left(T_{cr}^{x-j}, T_{cr}^{x-j} - T^\text{max}_\text{D}, \text{CollisionFree}\right)$
							\Comment{Using \eqref{eq:crit_time_diff_lower}}
						\Else
							\State $\underline{t}^{x-j}_0 = -\infty$
						\EndIf
						\State Append $[\underline{t}^{x-j}_0 + t^j_0 \;, \; \bar{t}^{x-j}_0 + t^j_0]$ to $avoidRange$						
			 		\EndIf
			 	\EndFor
			 	\State Sort $avoidRange$ by first column in ascending order
			 	\State $delay = 0$
			 	\For{each range $[lower, upper]$ in sorted $avoidRange$}
			 		\If{$delay < lower$}
			 		\State \textbf{break} \Comment{Found gap before this range}
			 	\ElsIf{$delay \in [lower, upper]$}
			 		\State $delay = upper$ \Comment{Move past this range}
			 	\EndIf
			 	\EndFor
			 	\State $timeAdded[x] = delay$			
		\EndFor		
		\end{algorithmic}
	\label{alg:timing_delay_cal}
\end{algorithm}

Considering the two main algorithms, the priority alocation algorithm has a time complexity of $O(n^2)$, whereas the delay-time calculation algorithm has a time complexity of $O(n^2 (\log n + \log T )))$ where $T$ denotes the ratio ($T^\text{max}_\text{D} / DT$) between maximum allowable delay ($T^\text{max}_\text{D}$) and step size parameter ($DT$). 
\par 

The proposed algorithm has several limitations: first, since it assumes drones travel in straight-line paths, it cannot solve for trajectories when fixed obstacles prevent direct paths between the starting position and the target position. Second, due to the topological sorting approach, the algorithm cannot calculate priorities when cyclic dependencies exist among multiple drones.

\section{Results}
\label{sec-results}

In this section, the effectiveness of the proposed algorithm is evaluated through simulations. First, the results of the proposed Time-Efficient Prioritised Scheduling  (TPS) algorithm are compared with those of the Coupling-Degree-Based Heuristic Prioritized Planning (CDH-PP) method \cite{Hui_Li_2019}, which also addresses a cooperative path planning problem for drone swarms. Both approaches assign a hierarchy among drones and perform sequential planning for individual drones to determine collision-free travel trajectories. However, CDH-PP can operate in an environments with fixed obstacles, whereas TPS assumes an obstacle-free environment. The algorithms are compared in terms of flock formation time, deviation from the optimal flock formation time, path length, and computation time in an obstacle-free setting. When comparing both algorithms, results for the same starting and target locations are considered, assuming similar drone capabilities. The CDH-PP algorithm is implemented based on the pseudocode and procedural descriptions provided, using the same algorithm-specific performance parameters. Subsequently, the proposed TPS algorithm is simulated with varying numbers of drones, including a scenario involving up to 5{,}000 drones, to evaluvate its scalability properties. All simulations were conducted using MATLAB R2020b on a workstation equipped with an AMD Ryzen Threadripper 3960X 24-Core Processor and 256 GB of RAM.

\subsection{Comparison with the CDH-PP Algorithm}

For all simulations, a given $n$ number of drones operate in a given 3D area. Drones begin their movement from randomly placed positions in a horizontal square XY plane centred around the starting position coordinates $[0,0,0]$ subject to parameters in Table \ref{tab:sample_metrics}. Note that it is assumed that the ground square plane is divided into grids with side length of 1 m. All start positions are located at grid points only. 

\begin{table}[H]
	\centering
	\caption{Common parameters}
	\begin{tabular}{|c|c|}
		\hline
		\textbf{Parameter Name} & \textbf{Parameter Value} \\
		\hline
		collision threshold ($R_{\text{col}}$) (m) & 1 \\
		minimum distance between two drones - starting position or target position (m) & 2 \\
		safety factor ($SF$)  & 1.5 \\
		Drone position density factor ($\delta$) & 10 \\
		maximum velocity ($V_{\text{max}}$) ($ms^{-1}$) & 20 \\
		maximum accelaration ($a_{\text{max}}$) ($ms^{-2}$) & 3 \\
		maximum decelaration ($d_{\text{max}}$) ($ms^{-2}$) & 3 \\
		\hline
	\end{tabular}
	\label{tab:sample_metrics}
\end{table}

When the available placement area is significantly larger than the number of deployed drones, the initial drone positioning becomes widely spaced, resulting in suboptimal space utilisation and overly simplified collision avoidance scenarios. Conversely, if the placement area is too small, achieving the minimum required distance between drones becomes impossible. Therefore, the drone positioning spacing factor should be realistic yet challenging. Due to the minimum required distance constraint between drones, each drone placement renders eight neighbouring positions ineligible. The current value of $10$ for $\delta$ results in a placement area that is demanding but achievable. Thus, the length of the side of the square $L_{sq}$ is given by ~\eqref{eq:squre_length}, which guarantees a minimum of ten possible positions per drone. 

\begin{equation}
	L_\text{sq} =  \left\lceil{\sqrt{n \times \delta}} \right\rceil
	\label{eq:squre_length}
\end{equation}

Similarly, target positions are also randomly assigned within a cube, with the furthest corner of the cube located at the coordinates of [200, 200, 200]. As with the starting positions, the volume in which drones can be placed should be realistic yet challenging. Due to spacing constraints, each drone placement renders twenty-six neighbouring positions ineligible. Note that, similar to the starting square, the cube is divided into smaller cubes with a side length of 1 m. Only grid points contain target positions. Considering all the requirements, the length of the side of the cube $L_{cu}$ is given by ~\eqref{eq:cube_length}, which guarantees a minimum of thirty positions per drone.

\begin{equation}
	L_\text{cu} =  \left\lceil \sqrt[3]{\text{n} \times \delta \times 3} \right\rceil
	\label{eq:cube_length}
\end{equation}

All starting and target positions are indexed from 1 to $n$, and each $i$-th drone moves from the $i$-th starting position to the $i$-th target position. The total flocking time ($t^n_\text{flock}$), is defined as the time elapsed between the moment the first drone starts moving, and the moment the last drone reaches its target position and stops moving.

Since the CDH-PP algorithm assumes constant velocity for all drones, the average velocity in the TPS algorithm simulation is used as the velocity for the CDH-PP algorithm. Two design parameters of the CDH-PP algorithm, namely the initial inflation factor $\epsilon$ is set to 1.5, while linear reduction $\Delta\epsilon$  is set to 0.1. These are the original parameters used by the authors of the CDH-PP algorithm.

\begin{table}[H]
	\caption{Comparison of TPS with CDH-PP}
	\centering
	\begin{tabularx}{\textwidth}{|>{\centering\arraybackslash}p{0.8cm}|>{\centering\arraybackslash}X|>{\centering\arraybackslash}p{1.8cm}|>{\centering\arraybackslash}p{1.8cm}|>{\centering\arraybackslash}p{1.8cm}|>{\centering\arraybackslash}X|>{\centering\arraybackslash}X|>{\centering\arraybackslash}p{1.3cm}|}
		\hline
		& \multicolumn{4}{c|}{\textbf{TPS}} & \multicolumn{2}{c|}{\textbf{CDH-PP}} & \textbf{Comparison} \\
		\cline{2-5} \cline{6-7}
		Number of Drones & Flock formation Time (s) \newline $t^n_\text{flock}$ & Average Delay (s) & Maximum Delay (s) & Calculation Time (s) \newline $T^n_\text{Cal}$ & Flock formation Time (s) \newline $t^n_\text{flock}$ & Calculation Time (s) \newline $T^n_\text{Cal}$ & Calculation Time Factor \\
		\hline
		10 & 22.509 ± 0.014 & 0.044 ± 0.004 & 0.213 ± 0.016 & 0.059 ± 0.001 & 22.627 ± 0.021 & 13.123 ± 0.277 & 221.119 \\
		15 & 22.608 ± 0.013 & 0.069 ± 0.005 & 0.339 ± 0.02 & 0.089 ± 0.002 & 22.794 ± 0.023 & 24.477 ± 0.578 & 274.431 \\
		20 & 22.625 ± 0.016 & 0.07 ± 0.004 & 0.405 ± 0.02 & 0.116 ± 0.001 & 22.781 ± 0.023 & 39.189 ± 1.519 & 337.797 \\	
		25 & 22.67 ± 0.015 & 0.092 ± 0.005 & 0.493 ± 0.022 & 0.15 ± 0.003 & 22.851 ± 0.023 & 64.037 ± 3.903 & 425.701 \\	
		30 & 22.712 ± 0.017 & 0.11 ± 0.005 & 0.584 ± 0.022 & 0.187 ± 0.003 & 22.934 ± 0.024 & 87.159 ± 7.346 & 467.059 \\
		\hline
	\end{tabularx}
	\label{tab:IS_vsCDHPP_results_CI}
\end{table}

Table \ref{tab:IS_vsCDHPP_results_CI} presents the summarised results for flock sizes ranging from 10 to 30 drones, in intervals of 5. In each replication, both TPS and CDH-PP algorithms use the same combination of randomly generated starting and target positions. For each drone count, 200 simulations are conducted, and the 95\% confidence intervals are presented in Table \ref{tab:IS_vsCDHPP_results_CI}.  The average delay is calculated by summing the starting delays of all $n$ drones and dividing this total by the number of drones ($(\sum_{i=1}^{n} t^i_0) / n$). The maximum delay is obtained by selecting the drone with the largest starting delay among all $n$ drones ($\max (t^i_0)$).

\begin{figure}[h]
	\centering
	\begin{tabular}{cc}
		\includegraphics[width=0.45\textwidth]{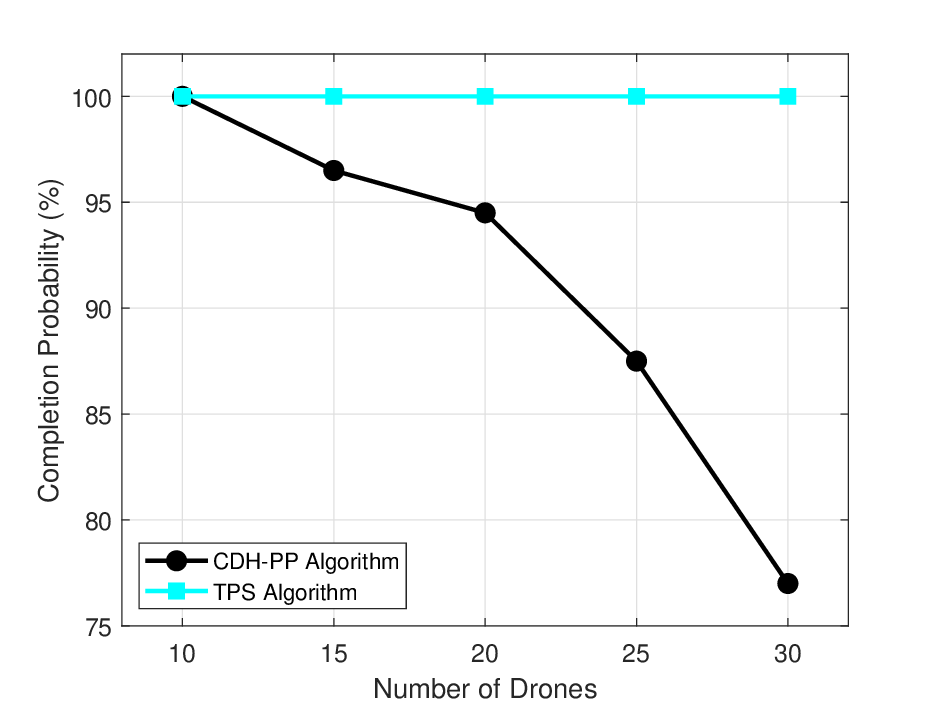} &
		\includegraphics[width=0.45\textwidth]{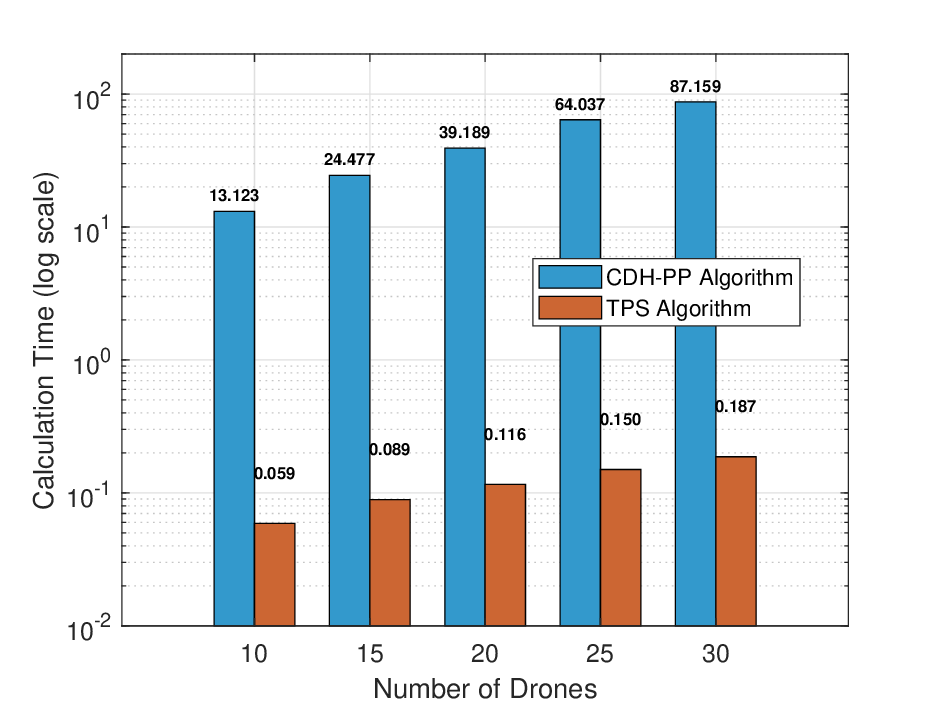} \\
		(a) Completion probability comparison & (b) Calculation time comparison \\[1em]
		\includegraphics[width=0.45\textwidth]{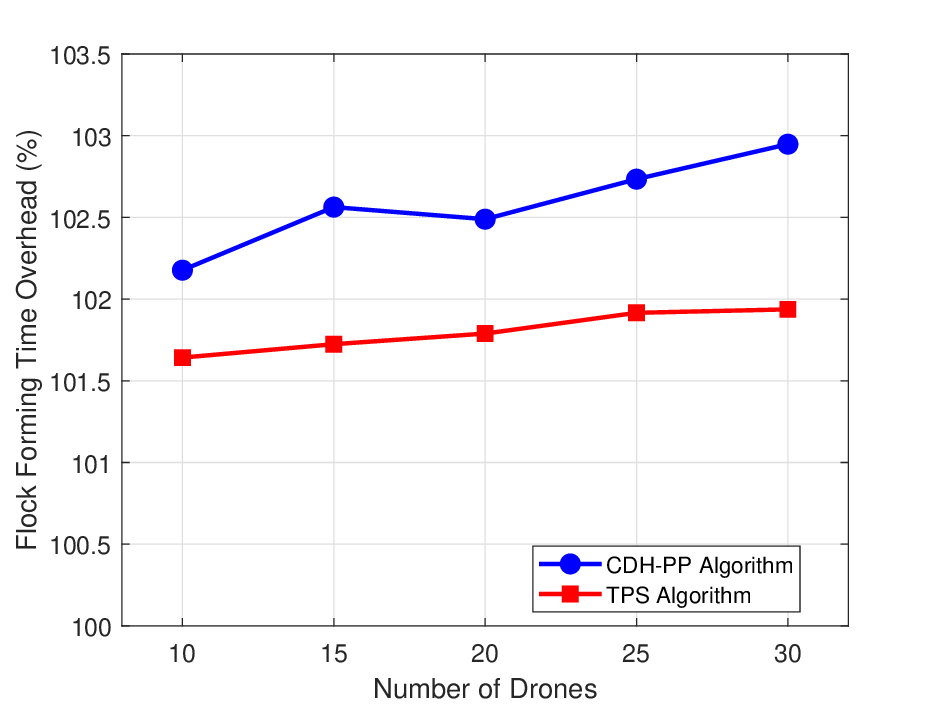} &
		\includegraphics[width=0.45\textwidth]{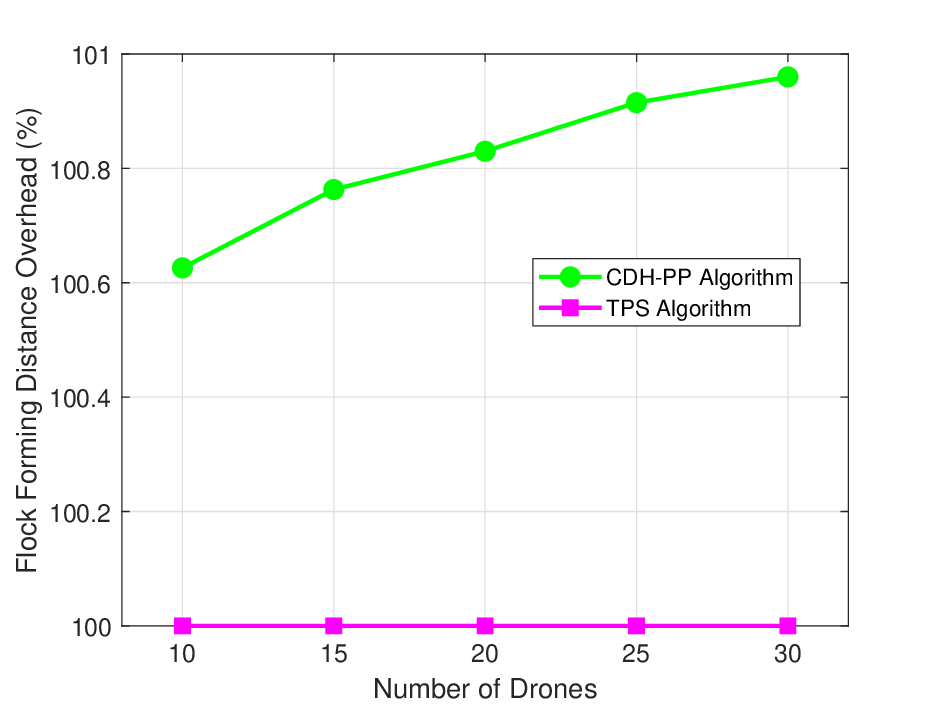} \\
		(c) Flock formation time overhead comparison & (d) Flock formation distance overhead comparison
	\end{tabular}
	\caption{Comparison between TPS with CDH-PP}
	\label{fig:four_figures_grid}
\end{figure}

Even though, Warnakulasooriya et al. \cite{Warnakulasooriya_2025} devised \eqref{eq:delta} to define a $\delta$ as a function of number of drones in the simulated flock ($n$) to calculate the optimal $\delta$ value for each $n$, for this comparison a fixed $\delta$ value of 10 is used. In this paper, the highest number of drones simulated is 30. When 30 is used in \eqref{eq:delta} as the number of drones, the minimum required $\delta$ value is 6.493, which is smaller than the $\delta$ value used for this simulation. Thus, a fixed $\delta$ value does not affect the performance of this research question.

\begin{equation}
	\delta(n) = 1.06 \times n ^ {0.5329}
	\label{eq:delta}
\end{equation}

Two key performance indicators are the average flock formation overhead time ($T^\text{OH}_\text{flock}$) and average flock formation overhead distance ($D^\text{OH}_\text{flock}$). Equations~\eqref{eq:overhead_time} and
\eqref{eq:overhead_dist} present calculations for $T^\text{OH}_\text{flock}$ and $D^\text{OH}_\text{flock}$ respectively, and the best possible result for each parameter being 100\%. In ~\eqref{eq:overhead_time}, denominator represents the travel time of the drone that travels the maximum distance, while numerator represents the overall flocking time from \eqref{eq:total_time_flock}. For a given configuration of start and target positions for $n$ drones, denominator value is fixed, while numerator value depends on the performance of the algorithm. 

\begin{equation}
	T^\text{OH}_\text{flock} = \frac{t^n_\text{flock}}{\max_{i=1}^{n} \left( t_{\text{travel}}^i \right)} 
	\label{eq:overhead_time}
\end{equation}

Similarly, in \eqref{eq:overhead_dist}, denominator represents the summation of distances that each drone can travel which is the straight-line distance from start to target positions. The numerator represents the summation of actual travel distances of each drone. Similar to \eqref{eq:overhead_time}, denominator value is fixed for a given configuration, while numerator value depends on the performance of the algorithm as actual travel path may increase if individual drones take longer routes to avoid collisions. 

\begin{equation}
	D^\text{OH}_\text{flock} = \frac{\sum_{i=1}^{n} D^i_\text{actual}}{\sum_{i=1}^{n} D^i_\text{straight-line}} 
	\label{eq:overhead_dist}
\end{equation}

When analysing performance of the TPS algorithm against the CDH-PP algorithm, first and most important characteristic is the completion probability comparison which is illustrated by Fig.~\ref{fig:four_figures_grid}(a). TPS provides a valid trajectory for all drones irrespective of the number of drones in the flock while CDH-PP gradually loses the ability to provide a collision-free valid trajectory for all $n$ drones when the flock size increases. When 30 drones are in the flock, collision-free solution rate of CDH-PP algorithm drops to 77\%. Similarly, as shown in Fig.~\ref{fig:four_figures_grid}(b), average calculation time for CDH-PP is much larger than the calculation times of TPS. 

When observing Fig.~\ref{fig:four_figures_grid}(c), both algorithms have a flock formation overhead time ($T^\text{OH}_\text{flock}$) slightly above 100\%. However, TPS algorithm performs marginally better than CDH-PP algorithm interms $T^\text{OH}_\text{flock}$. Finally, when considering flock formation overhead distance ($D^\text{OH}_\text{flock}$) as shown in Fig.~\ref{fig:four_figures_grid}(d), TPS algorithm consistently achieves the optimal travel distance while CDH-PP algorithm has a slight increase in $D^\text{OH}_\text{flock}$ as the number of drones in the flock increases.

%

\subsection{Scalability Performance of the TPS Algorithm}

In this paper the performance of TPS algorithm is evaluated starting from 50 drones to 5,000 drones in a flock to evaluate its scalability. The conditions for this evaluation are identical except for the target position coordinates and $\delta$ value. Due to the increase in the number of drones, the furthest corner of the cube for the target position is moved to coordinates [500, 500, 500]. Previously a fixed $\delta$ value of 10 was used. Warnakulasooriya et al. \cite{Warnakulasooriya_2025} defined $\delta$ as a function of $n$ to provide a 95\% cycle-free population using \eqref{eq:delta}. When the fixed $\delta$ value of 10 is substituted for \eqref{eq:delta}, corresponding $n$ value is 67 which is the maximum number of drones in a flock for the given $\delta$ value. However, since this simulation has flock sizes from 50 to 5,000, using a variable $\delta$ value becomes a necessity. For the previous evaluation where TPS and CDH-PP algorithms were compared, maximum flock count was 30. Thus for that experiment, 95\% cycle-free rate is not challenged. However, here the $\delta$ value must be adjusted as scalability is examined up to 5,000 drones. Also for each different number of drones, 200 simulation iterations are conducted and the 95\% confidence interval results are summarised in Table \ref{tab:results_summary}.

\begin{table}[H]
	\caption{Scalability performance of TPS}
	\centering
	\begin{tabularx}{\textwidth}{|>{\centering\arraybackslash}p{0.8cm}|>{\centering\arraybackslash}p{1.2cm}|>{\centering\arraybackslash}p{1.8cm}|>{\centering\arraybackslash}X|>{\centering\arraybackslash}X|>{\centering\arraybackslash}X|>{\centering\arraybackslash}X|>{\centering\arraybackslash}p{1.8cm}|}
		\hline
		Number of Drones &  $\delta$ Values & Cycle-Free Population Percentage (\%) &  Flock formation Time (s) \newline $t^n_\text{flock}$ & Mean Delay (s) & Maximum Delay (s) & Flock formation Overhead Time (\%) & Flock Formation Overhead Distance (\%) \\
		\hline
		50 & 8.5249 & 93\% & 44.127 ± 0.024 & 0.194 ± 0.007 & 0.938 ± 0.029 & 102.282\% & 100\% \\
		100 & 12.3341 & 88\% & 44.375 ± 0.022 & 0.262 ± 0.006 & 1.458 ± 0.028 & 102.317\% & 100\% \\
		250 & 20.0987 & 96.5\% & 44.957 ± 0.04 & 0.361 ± 0.006 & 2.684 ± 0.041 & 102.238\% & 100\% \\
		500 & 29.0795 & 94.5\% & 45.876 ± 0.042 & 0.507 ± 0.007 & 4.401 ± 0.055 & 102.087\% & 100\% \\
		1000 & 42.0732 & 95\% & 47.553 ± 0.041  & 0.734 ± 0.007 & 7.409 ± 0.079 & 101.885\% & 100\% \\
		2500 & 68.5596 & 96\% & 52.298 ± 0.061 & 1.548 ± 0.01 & 15.358 ± 0.118 & 101.902\% & 100\% \\
		5000 & 99.1944 & 96.5\% & 59.306 ± 0.081 & 2.846 ± 0.013 & 26.268 ± 0.168 & 101.918\% & 100\% \\
		\hline
	\end{tabularx}
	\label{tab:results_summary}
\end{table}

When analysing scalability properties from Table \ref{tab:results_summary}, it is evident that as the $\delta$ value increases, collision-free solution percentage close to the expected 95\%. In addition, flock formation time, mean delay, and maximum delay increases as the number of drones increase, while flock formation overhead time ($T^\text{OH}_\text{flock}$) remains around 102\%. Furthermore, the results confirm that if a collision-free solution exists, TPS algorithm can find feasible trajectories even as the number of drones increases in the flock up to 5,000.

\section{Discussion}
\label{sec-discussion}

When analysing the completion probability comparison under research question 1, TPS algorithm always provids a valid trajectory, whereas CDH-PP at times failed to provide a valid collision-free trajectory. The reason for this is the different approaches that two algorithms use to determine the hierarchy between drones as well as the approach both algorithms use to avoid potential collisions. In TPS algorithm, if a collision could occur between the drones, a single point is identified as the minimum distance point and this point is used to define the relationship between two drones. In contrast, in CDH-PP algorithm a count of individual trajectory points which are shorter than the collision radius is taken and stored at coupling degree matrix. A higher priority is given to drones with overall higher collision count with all the other drones. The issue with this method is two drones travelling close to each other will get a higher collision count resulting both these drones getting a higher priority. But in reality higher collision count was caused by a single other drone. This can lead to an incorrectly biased hierarchy among drones. The other issue contributing to the differences in success rate in providing a valid collision-free trajectory depends on the method each algorithm use to avoid collisions. TPS algorithm uses an initial delay-based approach and since blocking conditions are already analysed there will always be an initial delay for each drone that guarantees a collision-free trajectory. In CDH-PP trajectories are arranged sequentially starting from the highest priority drone. Anytime Repairing Sparse A* (AR-SAS) algorithm is used to generate individual trajectories. For each drone there is a time limit to generate a feasible trajectory without collisions. AR-SAS will plan paths as segments and at each segment, algorithm generates few different direction options but selects the option with the least cost. However, if there is a collision at the selected segment, AR-SAS algorithm will return to the segment starting position and select the next path option with the least cost. As the number of drones increases for drones with a lower priority, finding a collision-free trajectory using this iterative process becomes difficult inside the given time limit. This leads to some configurations without a feasible path. Results shown in Fig.~\ref{fig:four_figures_grid}(a) reflect these differences in the algorithms. \par 

Fig.~\ref{fig:four_figures_grid}(b) illustrates the calculation time differences between the two algorithms. Since paths are fixed in TPS algorithm, it first calculates the hierarchy and then followed by the computation of individual start time delays. But in CDH-PP algorithm, initially each drone calculates the best-case trajectory using AR-SAS, ignoring all other drones. Then these initial trajectories are used to compute the coupling degree matrix which is used to determine the hierarchy and sequence. Then except for the highest priority drone, all the other drones recalculate the trajectories considering higher priority drone trajectory points as obstacles. If a collision is detected, the drone under planning returns to the previous segment of its trajectory—up to the point where no collisions occurred—and regenerates a new trajectory from that point. Continuous path adjustments under an upper calculation time bound will consume substantial computational power depending on the individual trajectory calculation upper time limit and number of drones in the flock. This results in a significantly longer calculation times for CDH-PP algorithm compared with TPS. \par 

Flock forming time overhead ($T^\text{OH}_\text{flock}$) is presented in Fig.~\ref{fig:four_figures_grid}(c). As shown in the figure, both algorithms have a slight overhead that increases with number of drones. In TPS algorithm, this additional travel time is a result of individual start delays. In contrast, in CDH-PP algorithm overhead time arises due to the extra distance drones must travel along longer paths to avoid collisions. Since TPS has a slightly lower overhead compared with CDH-PP, it also outperforms CDH-PP in terms of time overhead. \par 

Flock forming distance overhead ($D^\text{OH}_\text{flock}$) is illustrated in Fig.~\ref{fig:four_figures_grid}(d). As all drones under TPS algorithm take the shortest possible path, there is no distance overhead from this method. However, CDH-PP takes longer paths to avoid collisions and, as a result, incurs a slight distance overhead. \par

Thus under all the criteria where both the algorithms were evaluated, the TPS algorithm outperforms the CDH-PP algorithm confirming its superior overall performance. \par 

When analysing scalability properties under research question 2, simulation results confirm that variable $\delta$ value as a function of $n$ guarantees a 95\% collision-free solution percentage. Flock forming time, mean delay and maximum delay values increse at a similar rate but most interestingly flock forming overhead time remains around 102\%. This behaviour can be explained as below. As the number of drones increases in the flock and as a result of the increase in $\delta$ value, drone starting position area and drone target position volume differ significantly between flock sizes. As the number of drones increases, travel distance difference between drone that travels the maximum distance and drone that travels the minimum distance increases substantially.Flock forming time depends on the maximum travel distance, while mean delay and maximum delay depend on the difference between the maximum and minimum travel distances. However, flock forming overhead time remains stable, as it is affected only by the drone density. Drone density remains consistent as a number of drone dependent variable $\delta$ value is utilised. Thus, we can conclude that flock forming overhead time depends solely on drone density, not on the number of drones in the flock itself.
\section{Conclusions}
\label{sec-conclusion}

This study addressed the challenge of achieving efficient and scalable initial flock formation for drones, a phase where existing algorithms often struggle with collision avoidance, travel efficiency, and scalability. To overcome these limitations, a time‑efficient prioritised scheduling (TPS) algorithm was proposed. The method establishes a hierarchy among drones considering potential collision risks and blocking likelihood, then assigns calculated delays to ensure collision‑free trajectories. By constraining paths to straight‑line segments, TPS minimises travel distances and reduces computational complexity. Simulation results confirm superior performance compared to the coupling‑degree‑based heuristic prioritized planning method (CDH‑PP) for flocks of up to 30 drones, and scalability was validated across larger swarms ranging from 50 to 5,000 drones.

These findings demonstrate that large‑scale drone initial flock formation can be achieved with both efficiency and reliability. The simplicity of TPS contributes to the broader body of research on swarm coordination by showing that collision‑free scheduling can be accomplished without heavy optimisation overhead. Practically, this opens opportunities for deploying drone swarms in real‑world applications such as drone light shows, aerial monitoring, logistics, and cooperative sensing, where rapid and reliable formation is critical. Despite its strengths, TPS has limitations. It cannot generate feasible solutions when permanent obstacles lie along straight‑line paths, and potential existance of a circular dependancy due to starting and target positions prevent valid delay assignments in some configurations. These constraints highlight the need for extensions to handle more complex environments. Existing limitations also point towards future research directions. Future work should integrate obstacle‑avoidance mechanisms into the scheduling framework, explore hybrid approaches that combine TPS with adaptive trajectory planning, and investigate robustness under dynamic conditions such as moving obstacles or communication delays. In addition, future research should also focus on finding viable trajectory planning method when circular dependancies exist.

By demonstrating scalability to thousands of drones while ensuring collision‑free and efficient initial formation, TPS represents a significant step toward practical and reliable drone coordination. This contribution lays the foundation for future advances in autonomous aerial systems, enabling drone swarms to operate safely and effectively in increasingly complex environments.

\section*{Acknowledgment}

S. Warnakulasooriya is supported by UC Doctoral Scholarship.



%

\bibliographystyle{ieeetr}
\bibliography{Refer.bib}

%

\begin{IEEEbiography}{Sujan Warnakulasooriya}
	Biography text here.
\end{IEEEbiography}

\begin{IEEEbiographynophoto}{Andreas Willig}
	Biography text here.
\end{IEEEbiographynophoto}


\begin{IEEEbiographynophoto}{Xiaobing Wu}
	Biography text here.
\end{IEEEbiographynophoto}




\end{document}